\documentclass[runningheads]{llncs}

\usepackage{eccv}

\usepackage{eccvabbrv}

\usepackage{graphicx}
\usepackage{booktabs}
\usepackage{comment}
\usepackage{multirow}
\usepackage{wrapfig} 
\usepackage{makecell}
\usepackage[accsupp]{axessibility}  

\usepackage{hyperref}

\usepackage{orcidlink}

\begin{document}

\title{Cascaded Diffusion Framework for Probabilistic Coarse-to-Fine Hand Pose Estimation}

\titlerunning{Coarse-to-Fine Diffusion for 3D Hand Pose Estimation}

\author{Taeyun Woo \and
Jinah Park \and
Tae-Kyun Kim}

\authorrunning{Woo et al.}

\institute{KAIST\\
\email{\{taeyun.woo, jinahpark, kimtaekyun\}@kaist.ac.kr}}

\maketitle

\begin{figure}[t]
  \centering
  \includegraphics[width=\linewidth]{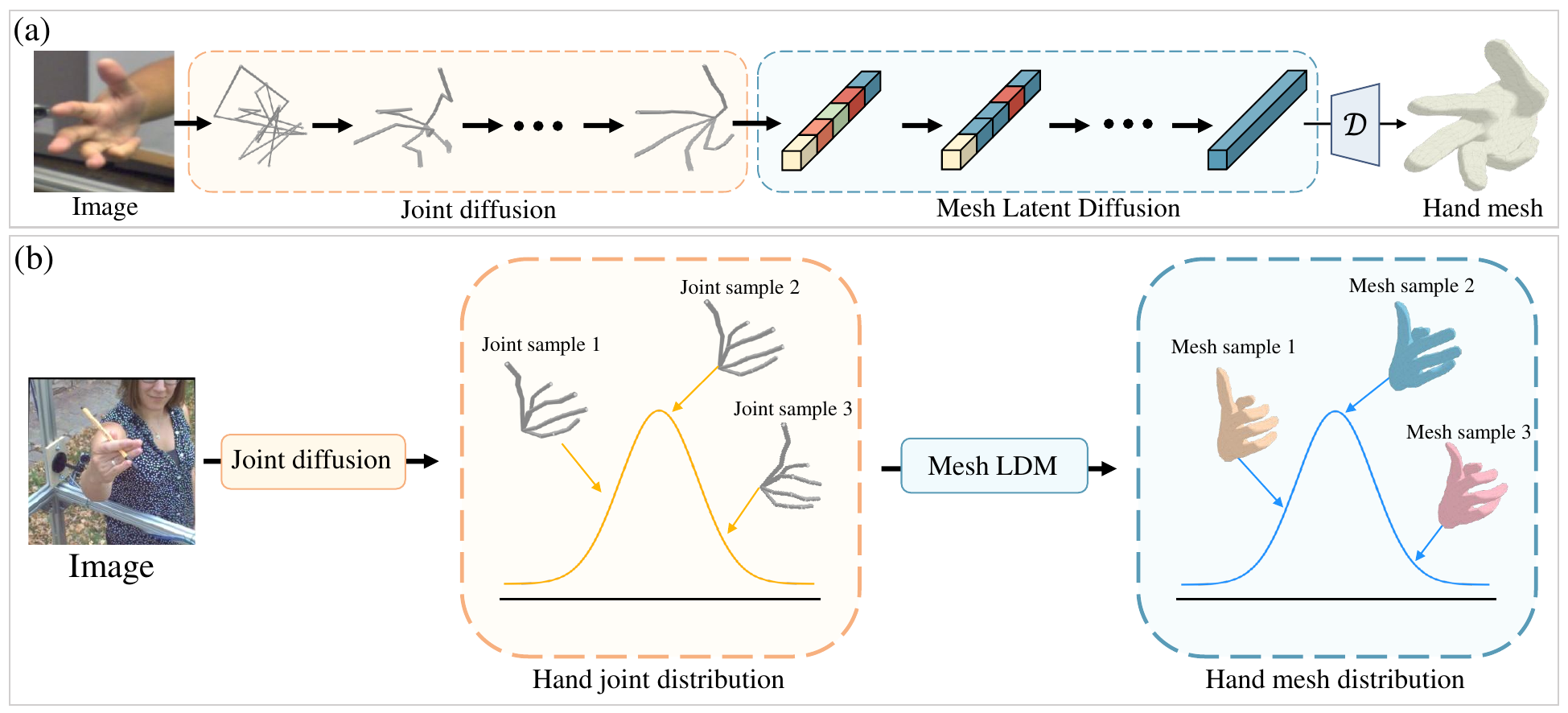}
  \caption{\textbf{A cascaded diffusion model for hand pose estimation.}  (a) The joint diffusion model first denoises noise into 3D hand keypoints, and the Mesh Latent Diffusion Model (Mesh LDM) then reconstructs the hand mesh from them. (b) Rather than conditioning on a single keypoint estimate, the Mesh LDM is trained on diverse joint samples drawn from a pose distribution, so it learns distribution-aware joint–mesh relationships.}

  \label{fig:teaser}
\end{figure}

\begin{abstract}
Deterministic models for 3D hand pose reconstruction, whe- ther single-staged or cascaded, struggle with pose ambiguities caused by self-occlusions and complex hand articulations. Existing cascaded approaches progressively refine pose predictions in a coarse-to-fine manner, but their deterministic nature prevents them from modeling pose uncertainty. Conversely, recent probabilistic methods capture pose distributions but are confined to single-stage estimation, often yielding inaccurate 3D reconstructions without refinement. To address these limitations, we propose a coarse-to-fine cascaded diffusion framework that combines probabilistic modeling with cascaded refinement. The first stage is a joint diffusion model that samples diverse 3D joint hypotheses, and the second stage is a Mesh Latent Diffusion Model (Mesh LDM) that reconstructs a 3D hand mesh conditioned on a joint sample. Our key idea is to use these diverse hypotheses as a training signal rather than as final outputs, so that the Mesh LDM learns distribution-aware joint–mesh relationships and becomes robust to the variation in coarse predictions. Extensive ablations validate the necessity of the cascaded design and the choice of latent-space diffusion. Experiments on FreiHAND, HO3Dv2, and DexYCB show that our method achieves state-of-the-art performance and remains stable under occlusion and pose ambiguity.

\end{abstract}


\section{Introduction}
\label{sec:intro}

3D hand pose estimation (HPE) is an important research area for emerging applications such as VR/AR \cite{guleryuz2018fast, shi2022grasping}, robotics \cite{li2019vision, handa2020dexpilot, lopez2023dexterous}, and human-computer interaction \cite{sharma2019grasping, sharma2021solofinger}. HPE aims to predict hand poses from images, and numerous approaches \cite{moon2020i2l, choi2020pose2mesh, chen2021i2uv, lin2021end, tang2021towards, lin2021mesh, chen2022mobrecon, li2024hhmr, dong2024hamba, potamias2024wilor, chen2025handos} have achieved promising results. Despite recent progresses, HPE remains a challenging task due to self-occlusions, complex hand articulation, and diverse hand shapes.

Traditional HPE approaches~\cite{hasson2019learning, zimmermann2019freihand, lin2021end} typically adopt a single-stage pipeline that directly regresses 3D joint positions or parameterized hand models such as MANO~\cite{romero2017embodied} from input images. However, by estimating 3D hand poses in a single step, these methods lack a refinement process that can progressively resolve pose ambiguities. As a result, learning the highly non-linear mapping from 2D observations to 3D hand configurations becomes particularly challenging under occlusions and complex articulations.

To address these issues, cascaded architectures~\cite{moon2020i2l, choi2020pose2mesh, tang2021towards, chen2022mobrecon, potamias2024wilor} have been proposed, where an initial hand pose estimation (\eg, joint positions or MANO parameters) is progressively refined through subsequent stages. By decoupling HPE into coarse estimation and fine-level refinement, cascaded models have shown improved performance compared to single-stage models. However, these methods remain deterministic, producing only a single prediction and failing to capture the uncertainty and diversity of valid hand poses, which limits their ability to handle the pose ambiguities.

Recently, denoising diffusion models~\cite{ho2020denoising, rombach2022high} emerged as powerful generative frameworks for modeling complex data distributions. They have shown remarkable success in image generation~\cite{ramesh2021zero}, 3D object generation~\cite{koo2023salad}, and human motion synthesis~\cite{tevet2023human}. Beyond generative tasks, diffusion models also show potential in human pose estimation \cite{shan2023diffusion, gong2023diffpose, feng2023diffpose, holmquist2023diffpose} and hand pose estimation \cite{ivashechkin2023denoising, cheng2024handdiff, li2024hhmr}, leveraging the generative ability to model pose distributions. These models are capable of sampling diverse pose hypotheses from complex distributions, however, existing methods adopt the single-stage design, which limits their ability to refine noisy predictions or recover fine-grained details.

In this paper, we propose a novel cascaded diffusion framework for 3D hand pose estimation (\cref{fig:teaser}) that combines the coarse-to-fine cascaded framework with the probabilistic modeling power of diffusion models. Specifically, our method consists of two stages: a joint diffusion model that denoises 3D keypoints conditioned on 2D input, and a mesh latent diffusion model (Mesh LDM) that reconstructs the 3D mesh latent vector conditioned on the denoised joint and image features. Our key idea is to use the diverse joint hypotheses from the first stage as a condition for training the second stage, rather than as final outputs. This allows the second stage to learn from the distribution of plausible joints rather than a single deterministic estimate, improving robustness when the coarse prediction is uncertain.

\begin{wrapfigure}{r}{0.55\linewidth}  
   \vspace{-0.5cm}
  \centering
   \includegraphics[width=\linewidth]{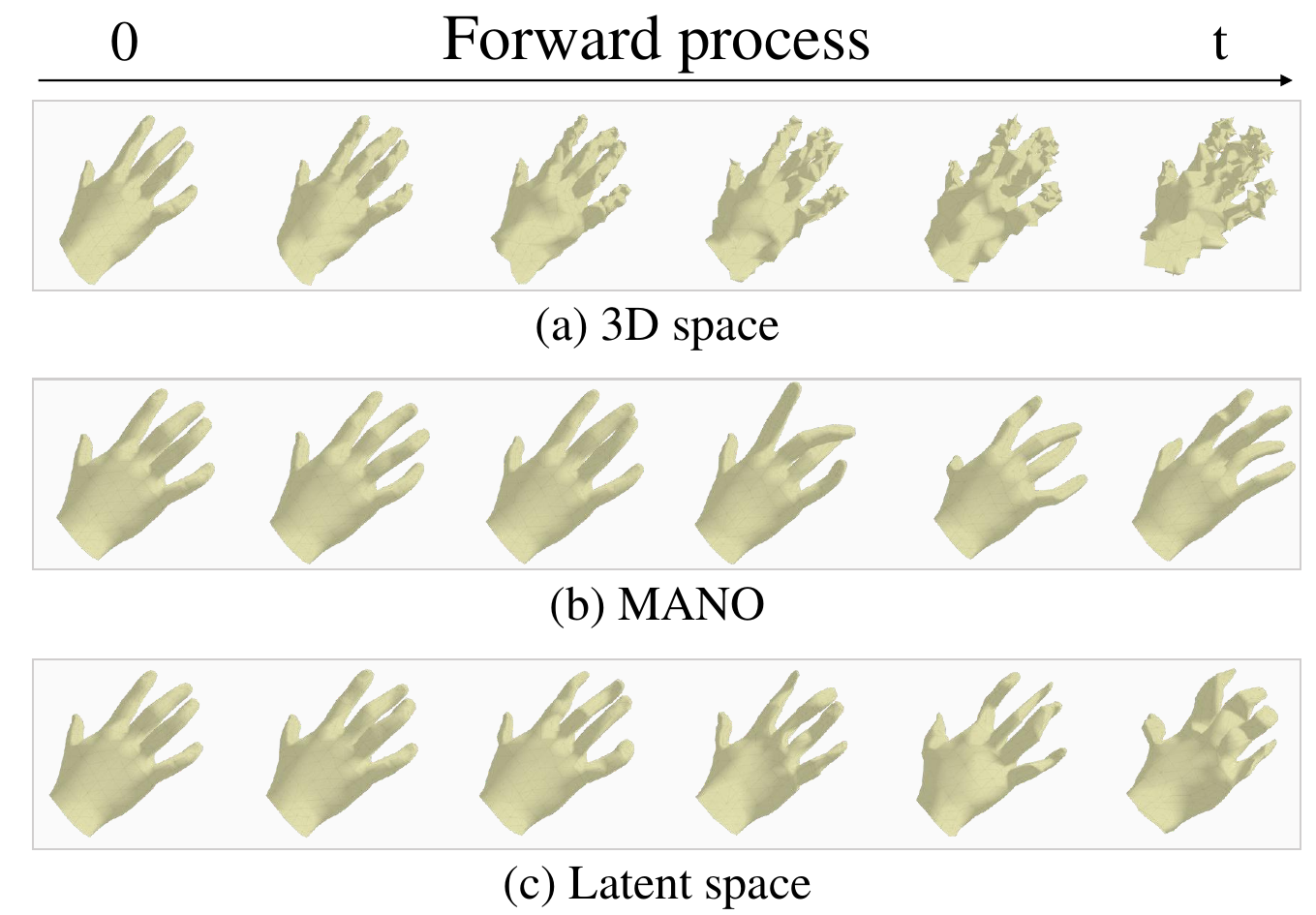}
   \caption{\textbf{Forward diffusion process in different spaces.} The figure shows how the noise progressively changes the 3D hand mesh across different representations.
   }
   \label{fig:forward_process}
   \vspace{-0.5cm}
\end{wrapfigure}

Unlike previous diffusion-based methods that operate in 3D space~\cite{li2024hhmr} or MANO space~\cite{cho2023generative}, our approach performs diffusion in a latent space. As illustrated in \cref{fig:forward_process}, diffusion in (a) 3D space often leads to the loss of surface geometry and pose structure. By contrast, (b) MANO space leverages a hand shape prior with predefined hand parts and can better preserve hand shapes, but it still tends to lose joint articulation fidelity. In (c), diffusion in the latent space preserves both pose structure and surface details, resulting in more expressive and realistic hand representations.

Furthermore, since the mesh diffusion model is conditioned on a joint sample drawn from the joint diffusion model, it learns mesh distributions over a range of plausible poses rather than a single set of deterministic keypoints. This design allows the model to capture distribution-aware joint-mesh relationships, improving robustness and performance in challenging scenarios.

Our contributions are summarized as follows: 
\begin{itemize}
\item[$\bullet$] We propose a cascaded diffusion framework for hand pose estimation that couples a probabilistic joint stage with a mesh refinement stage in a coarse-to-fine manner.
\item[$\bullet$] By conditioning the mesh diffusion model on a joint sample from diverse joint hypotheses generated from joint diffusion model, our method learns distribution-aware joint–mesh relationships. This improves accuracy and robustness to pose ambiguity.
\item[$\bullet$] Our method achieves state-of-the-art performance on DexYCB \cite{chao2021dexycb} and competitive results on FreiHAND \cite{zimmermann2019freihand} and HO3Dv2 \cite{hampali2020honnotate} with extensive experiments validating its effectiveness.
\end{itemize}

\section{Related works}
\subsection{3D Hand Pose Estimation}
3D hand pose estimation (HPE) has been extensively studied, with numerous approaches proposed over the past decade \cite{baek2019pushing, boukhayma20193d, zhang2019end, ge20193d, kulon2020weakly, baek2020weakly, armagan2020measuring, moon2020i2l, choi2020pose2mesh, chen2021i2uv, lin2021end, tang2021towards, lin2021mesh, chen2022mobrecon, lee2023fourierhandflow, lee2023im2hands, li2024hhmr, dong2024hamba, potamias2024wilor, cho2024dense, chen2025handos, lee2025rewind, woo2024hand}. One common paradigm in this field \cite{hasson2019learning, baek2019pushing, boukhayma20193d, zhang2019end, chen2022mobrecon, pavlakos2024reconstructing, dong2024hamba, potamias2024wilor} involves predicting a parameterized hand model, such as MANO \cite{romero2017embodied}, from an input image. These methods leverage prior knowledge of the hand model for robust estimation but are inherently limited by the predefined hand shape and lack fine-grained pose variations. Alternatively, other approaches \cite{ge20193d, kulon2020weakly, choi2020pose2mesh, moon2020i2l, lin2021end, li2024hhmr} predict hand keypoints and mesh vertices directly without predefined priors, offering greater flexibility. However, these single stage approaches often suffer from geometric inconsistencies and noisy predictions.

To improve robustness and accuracy, cascaded models \cite{choi2020pose2mesh, moon2020i2l, chen2022mobrecon, potamias2024wilor} decompose the HPE pipeline into a coarse-to-fine two-stage model. The first stage predicts an initial pose (\eg, 3D keypoints or MANO parameters), while the second stage refines the prediction by incorporating additional details. However, existing deterministic cascaded models fail to model pose uncertainty, particularly in occluded or ambiguous scenarios. This limitation motivates the exploration of stochastic approaches, such as diffusion models, for HPE.

\subsection{Diffusion-Based Pose Estimation}
Denoising diffusion models \cite{ho2020denoising, rombach2022high} have demonstrated remarkable success in generative tasks, such as image synthesis \cite{ramesh2021zero, rombach2022high}, 3D object generation \cite{koo2023salad, lee2024interhandgen}, and human motion synthesis \cite{tevet2023human, liang2024intergen}. Their ability to capture complex distributions has recently been explored in human pose estimation \cite{shan2023diffusion, gong2023diffpose, feng2023diffpose, holmquist2023diffpose, cho2023generative, dposer, foo2023distribution}. However, their application to hand poses remains largely unexplored, with only a few recent works \cite{ivashechkin2023denoising, cheng2024handdiff, li2024hhmr}. 

Existing diffusion-based hand pose estimating methods typically adopt a single-stage approach, directly estimating either hand keypoints \cite{ivashechkin2023denoising, cheng2024handdiff} or meshes \cite{li2024hhmr}. While diffusion models effectively model complex distributions, these methods lack a structured refinement process, limiting their robustness under occlusions and complex articulations. This motivates our cascaded diffusion framework, which introduces a coarse-to-fine refinement process into diffusion-based hand estimation.

\subsection{Cascaded Diffusion Models} While traditional diffusion models operate in a single-stage fashion, recent works have introduced cascaded diffusion frameworks across tasks such as image generation \cite{ho2022cascaded, shabani2024visual, li2025likelihood}, 3D object generation \cite{koo2023salad, lee2024interhandgen}, and video synthesis \cite{zhang2023i2vgen, jain2024video}. These models progressively refine outputs by conditioning each stage on the results of the previous stage.

In 3D domains, InterHandGen \cite{lee2024interhandgen} generates two-hands poses in two steps, and SALAD \cite{koo2023salad} decomposes 3D object synthesis into multiple representations. While InterHandGen generates interaction scenarios and SALAD focuses on
generating static objects, our method estimates articulated 3D hand poses from an image. Unlike these generation tasks, where diverse outputs are the goal, our cascade serves accuracy: the diverse hypotheses of the first stage are a means to train a more robust refinement stage.

The key advantage of cascaded diffusion in regression tasks is that the second stage is conditioned on a sample drawn from a distribution of plausible outputs, rather than a single deterministic one. This enables greater robustness by leveraging the stochastic nature of diffusion models, allowing better handling of ambiguities and pose uncertainty. Our proposed cascaded diffusion framework extends this idea to hand pose estimation, ensuring that the generated samples represent the distribution of 3D hand poses.

\begin{figure*}[htb]
  \centering
   \includegraphics[width=0.95\linewidth]{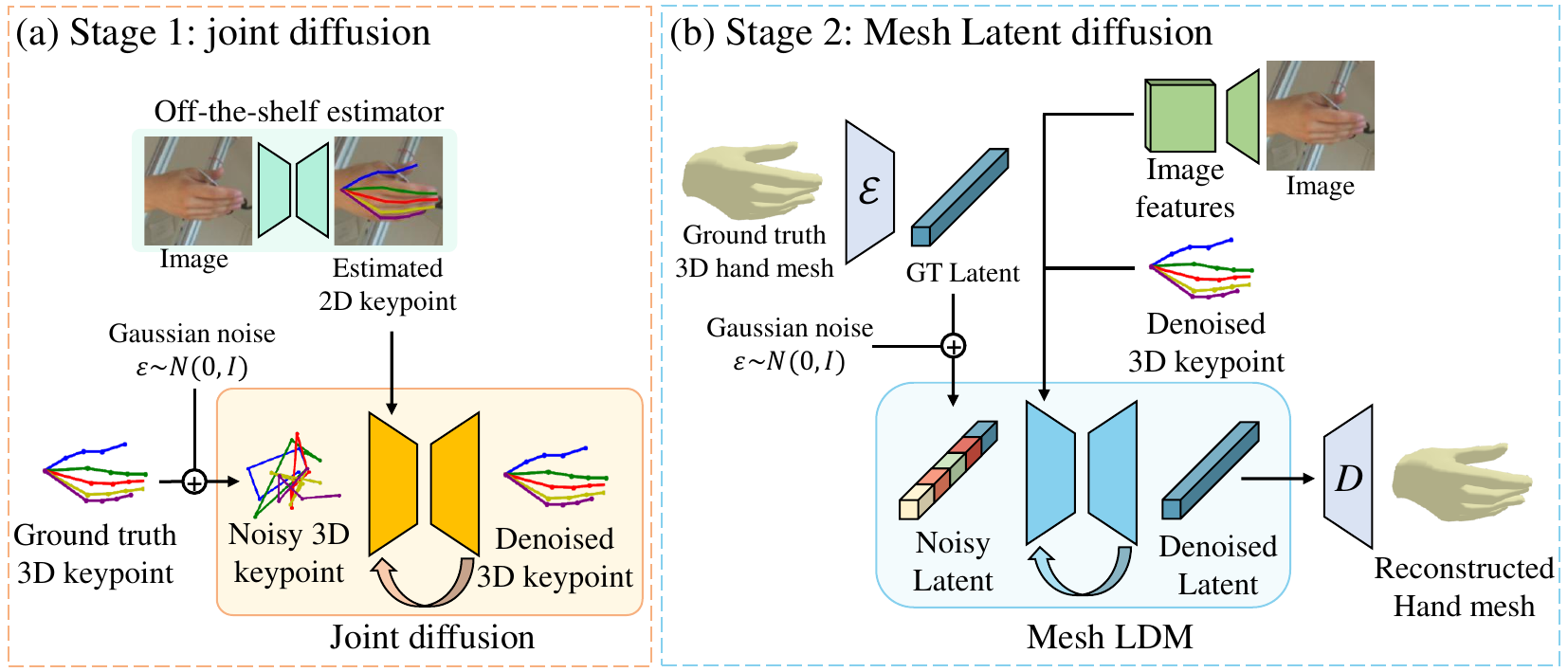}
   \vspace{-0.2cm}
   \caption{\textbf{Overview of the proposed cascaded diffusion model.} (a) The joint diffusion model lifts 2D keypoints from an off-the-shelf estimator into 3D joints. (b) With the joint model fixed, the Mesh LDM denoises the hand-mesh latent, conditioned on a denoised 3D joint and image features. The latent is decoded into the final mesh by the pre-trained Mesh AE decoder.}
   \label{fig:main}
   \vspace{-0.3cm}
\end{figure*}

\section{Method} \label{sec:method}
As shown in \cref{fig:main}, we propose a coarse-to-fine cascaded diffusion framework for 3D hand pose estimation. Our model consists of two stages: a joint diffusion model that estimates 3D hand joints from 2D keypoints, and a mesh latent diffusion model (Mesh LDM) that reconstructs the 3D hand mesh from the denoised joint and image features. This probabilistic coarse-to-fine design allows the model to represent multiple plausible joint hypotheses  instead of a single deterministic prediction, and refine the mesh conditioned on the learned joint distribution.

\subsection{Background: Denoising Diffusion Model}
\label{sec:background}

Denoising diffusion models~\cite{ho2020denoising,rombach2022high} generate samples through a forward process that gradually corrupts data with Gaussian noise and a reverse process that learns to denoise it. The forward process follows a Markov chain $q(\mathbf{x}_t \mid \mathbf{x}_{t-1}) = \mathcal{N}(\mathbf{x}_t; \sqrt{1-\beta_t}\,\mathbf{x}_{t-1}, \beta_t \mathbf{I})$, where $\beta_t$ is the noise variance at step $t$, and the reverse process $p_\theta(\mathbf{x}_{t-1}\mid\mathbf{x}_t)$ is learned by the model. Following previous pose estimation approaches~\cite{shafir2023human, tevet2023human, li2024hhmr}, we adopt the $\mathbf{x}_0$-parameterization, training the model to directly reconstruct the clean data $\mathbf{x}_0$ rather than the added noise with following denoising loss: $\mathcal{L}_{DDPM} = \mathbb{E}_{\mathbf{x}_0, t} \left[ \| \mathbf{x}_0 - \hat{\mathbf{x}}_0 \|^2 \right].$

\subsection{Stage 1: Joint Diffusion Model}
\label{sec:stage1}
The joint diffusion model lifts 2D hand keypoints into 3D joints. It adapts D3DP~\cite{shan2023diffusion}, originally proposed for 2D-to-3D human pose sequence uplifting, to the hand domain. Given a noisy joint $J_t$, the model denoises it into a clean 3D joint $\hat{J}_0$, conditioned on 2D keypoints $c_{2D}$. 

\vspace{2pt}
\noindent\textbf{Training.}
The model takes three inputs: a timestep $t \sim \mathcal{U}(0,T)$, 2D keypoints $c_{2D}$ from an off-the-shelf estimator~\cite{potamias2024wilor}, and a noisy joint $J_t$. The joint diffusion model is trained to directly reconstruct the clean joint by minimizing the diffusion loss $\lVert J_0 - \hat{J}_0 \rVert^2$.
 
\vspace{2pt}
\noindent\textbf{Inference.}
Starting from Gaussian noise $\epsilon \sim \mathcal{N}(0,\mathbf{I})$, the model progressively denoises it into a 3D joint $\hat{J}_0$ conditioned on $c_{2D}$. Each generated joint serves as a condition for the Mesh LDM.

\subsection{Stage 2: Mesh Latent Diffusion Model}
\label{sec:stage2}

\vspace{2pt}
\noindent\textbf{Mesh AutoEncoder.}
To embed the hand mesh into a latent space, we train a Mesh AutoEncoder (Mesh AE) based on SpiralNet++~\cite{gong2019spiralnet}. The encoder $\mathcal{E}$ maps a hand mesh $V \in \mathbb{R}^{778 \times 3}$ to a latent vector $x \in \mathbb{R}^{168}$, and the decoder $\mathcal{D}$ reconstructs the mesh from it. Mesh AE is trained with vertex and joint reconstruction losses together with
a KL term that regularizes the latent toward a Gaussian. Here the joint is obtained from the mesh via the MANO joint regressor~\cite{romero2017embodied}.

\vspace{2pt}
\noindent\textbf{Mesh Latent Diffusion.}
Mesh LDM $p_\phi$ denoises a noisy latent $x_t$ into the clean latent $x_0$, conditioned on the joint sample $\hat{J}_0$ from the joint diffusion model and on image features $\mathcal{I}$. The denoised latent is then decoded into the hand mesh, $\hat{V}_0 = \mathcal{D}(\hat{x}_0)$.

Mesh LDM is built on the DiT framework~\cite{peebles2023scalable} and operates on a sequence of $21$ tokens, one per hand joint. To form the input, the noisy latent is broadcast across the $21$ positions and concatenated along the channel axis with the denoised joint $\hat{J}_0$, so that every token carries the full mesh latent together with the coordinate of its corresponding joint. The resulting sequence is projected to the hidden dimension of the transformer.

\begin{wrapfigure}{r}{0.6\linewidth}  
   \vspace{-0.5cm}
  \centering
   \includegraphics[width=\linewidth]{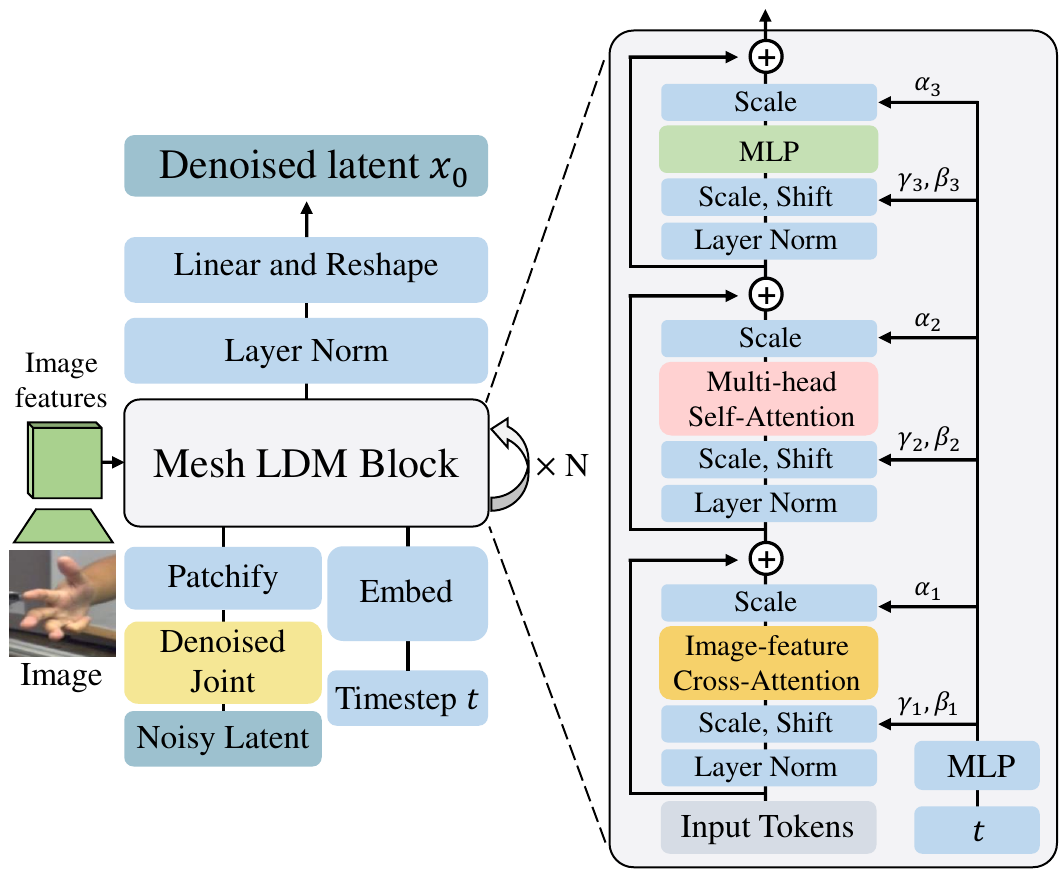}
   \caption{\textbf{Mesh LDM architecture.} The latent input and denoised joint are processed through DiT blocks  \cite{peebles2023scalable} with cross-attention to image features.}
   \label{fig:block}
   \vspace{-0.4cm}
\end{wrapfigure}

Image features are extracted as a four-level feature pyramid, and each level is projected to the hidden dimension and flattened into a set of spatial tokens. Within each block, the tokens are first refined by self-attention; they then attend to the four feature levels through parallel cross-attentions, whose outputs are concatenated and fused by an MLP. The diffusion timestep modulates every block through adaptive layer normalization~\cite{perez2018film}. After the final block, the tokens are projected and flattened back to the denoised latent $\hat{x}_0$. The overall structure is illustrated in \cref{fig:block}.

\subsection{Cascaded Diffusion Framework}
\label{sec:cascade}
Our cascaded framework couples the two stages so that mesh reconstruction is guided by the joint distribution rather than a single deterministic pose. Instead of regressing the mesh directly from the image, we decompose the task into 3D joint estimation followed by mesh reconstruction: the joint diffusion model
provides diverse joint hypotheses, and the Mesh LDM reconstructs a mesh conditioned on them.
 
\vspace{2pt}
\noindent\textbf{Training.}
We first train the joint diffusion model $p_\theta$ to lift 2D keypoints into 3D joints. Its weights are then fixed, and the Mesh LDM is trained to reconstruct a mesh latent conditioned on a single joint sample drawn from $p_\theta$, together with image features $\mathcal{I}$. Since $p_\theta$ produces different hypotheses at each step, the Mesh LDM is exposed to a range of plausible joint samples over training. This lets it learn a joint--mesh mapping that is robust to the variation in coarse predictions, rather than overfitting to a single deterministic input.
 
\vspace{2pt}
\noindent\textbf{Inference.}
At inference, the joint diffusion model generates multiple joint hypotheses, which are averaged into a stable estimate. Conditioned on this aggregated joint and image features, the Mesh LDM generates several latent samples that are averaged and decoded into the final mesh. Aggregating at both stages turns the diversity seen during training into a stable point estimate at test time.

\subsection{Loss Functions}
\label{sec:loss}
 
We employ three loss terms: a diffusion loss, a mesh vertex loss, and a joint loss. Since the joint diffusion model and Mesh AE are already trained, their parameters are fixed while training the Mesh LDM.
 
The diffusion loss supervises the Mesh LDM to reconstruct the clean latent under the $\mathbf{x}_0$-parameterization, $\mathcal{L}_{DDPM}=\lVert x_0-\hat{x}_0\rVert^2$. The vertex and joint losses enforce accurate geometry and pose consistency:
\begin{equation}
\mathcal{L}_{V}=\big\lVert V-\hat{V}\big\rVert_{1},
\qquad
\mathcal{L}_{J}=\big\lVert J-\mathcal{J}\hat{V}\big\rVert_{1},
\end{equation}
where $\mathcal{J}$  is the joint regressor that extracts 3D joints from the mesh
vertices. The total training loss is
\begin{equation}
\mathcal{L}=\lambda_{DDPM}\mathcal{L}_{DDPM}
+\lambda_{V}\mathcal{L}_{V}
+\lambda_{J}\mathcal{L}_{J},
\end{equation}
where we set $\lambda_{DDPM}=1$, $\lambda_{V}=10$, and $\lambda_{J}=5$.

\section{Experiments} 
\begin{figure*}[h]
   \vspace{-0.4cm}
  \centering
   \includegraphics[width=\linewidth]{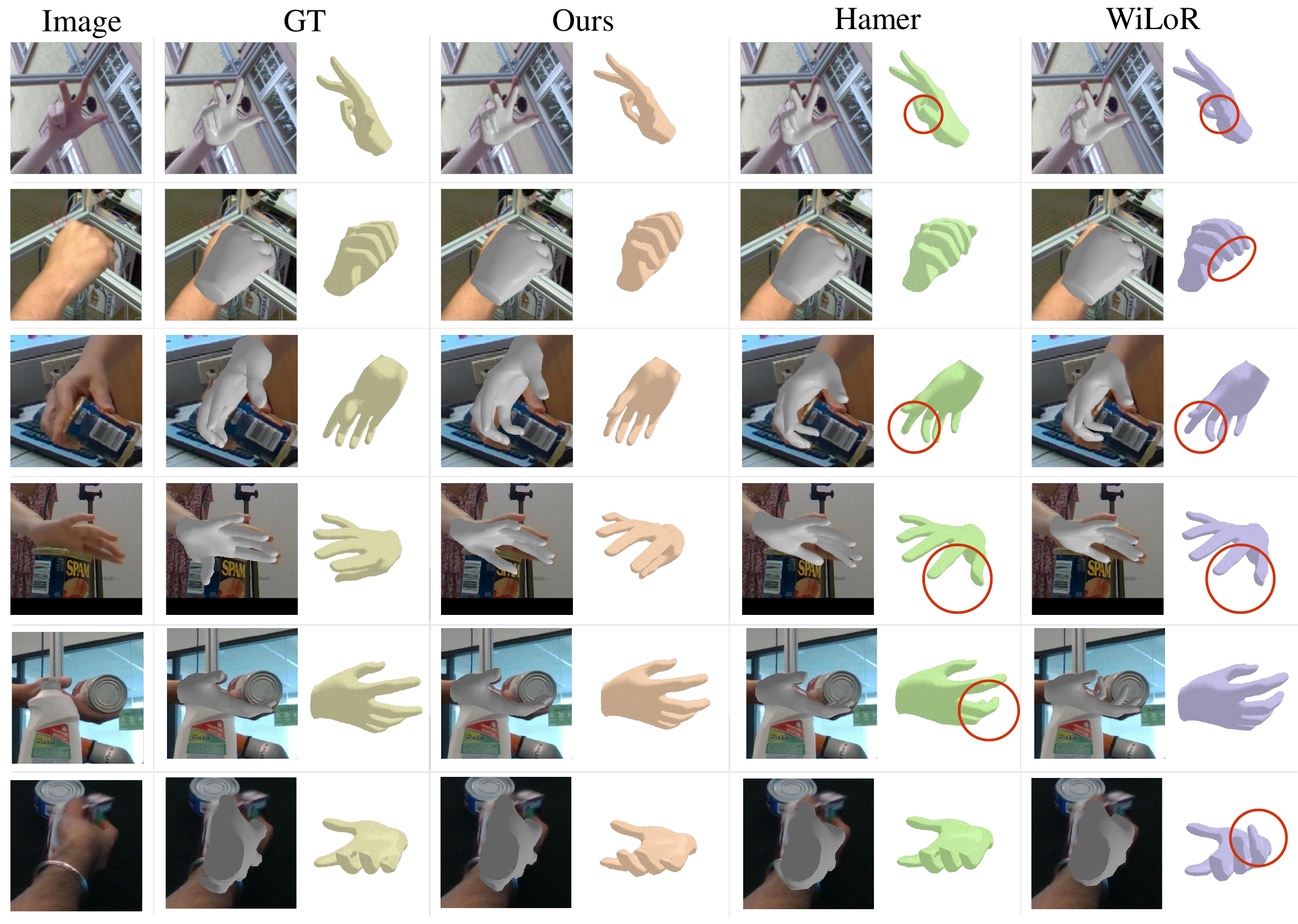}
   \caption{\textbf{Qualitative results} on FreiHAND \cite{zimmermann2019freihand}, HO3Dv2 \cite{hampali2020honnotate} and DexYCB \cite{chao2021dexycb}.}
   \label{fig:main_qualitative}
   \vspace{-0.5cm}
\end{figure*}

\subsection{Experimental settings}
\noindent\textbf{Implementation details.}
Our framework is implemented in PyTorch. The joint diffusion model is based on D3DP~\cite{shan2023diffusion}, which uplifts human pose sequences to 3D using the MixSTE backbone~\cite{zhang2022mixste}; we set the input sequence length to 1 for single-frame estimation. The Mesh AutoEncoder is trained on each dataset separately. Both the joint diffusion model and Mesh LDM use 1000 denoising steps for training, and DDIM sampling~\cite{song2020denoising} with a step size of 10 at inference. Further details are provided in the Supplementary Material.

\vspace{2pt}
\noindent\textbf{Datasets.}
We evaluate on three widely used hand pose benchmarks that span increasing levels of difficulty: FreiHAND~\cite{zimmermann2019freihand}, HO3Dv2~\cite{hampali2020honnotate}, and DexYCB~\cite{chao2021dexycb}. FreiHAND is a single-hand dataset with 133K training and 3.9K evaluation images, containing relatively little occlusion. HO3Dv2 and DexYCB are hand-object interaction datasets that involve object-induced occlusion: HO3Dv2 provides 66K training and 11K test images, while DexYCB is substantially larger, with 465K training and 116K test images. We follow the official split for all three datasets.

\vspace{2pt}
\noindent\textbf{Training details.}
We train our cascaded diffusion model separately on each dataset using the AdamW optimizer~\cite{loshchilov2017decoupled} on a single NVIDIA RTX 4090 GPU with a mini-batch size of 32. The joint diffusion model is trained for 250K iterations and the Mesh LDM for 100K iterations. The initial learning rate is 1e-4 and decays by a factor of 0.9 every 5K iterations under a step-based scheduler.

\vspace{2pt}
\noindent\textbf{Evaluation metrics.}
Following standard evaluation protocols, we assess performances using Procrustes Aligned Mean Per Joint Position Error (P-MPJPE) and Procrustes Aligned Mean Per Vertex Position Error (P-MPVPE). Additionally, we report the fraction of poses with errors below 5mm (F@5) and 15mm (F@15). For DexYCB we additionally report the Area Under the Curve (AUC) for 3D hand joints following HandOS~\cite{chen2025handos}. 

\begin{table}[t]
\small
\centering
\begin{tabular}{@{}l@{\hspace{3pt}}c@{\hspace{3pt}}c@{\hspace{3pt}}c@{\hspace{3pt}}c@{}}
\toprule
Method & P-MPJPE ↓ & P-MPVPE ↓ & F@5 ↑ & F@15 ↑ \\
\midrule
I2L-MeshNet \cite{moon2020i2l} & 7.4 & 7.6 & 0.681 & 0.973 \\
Pose2Mesh \cite{choi2020pose2mesh} & 7.7 & 7.8 & 0.674 & 0.969 \\
I2UV-HandNet \cite{chen2021i2uv} & 6.7 & 6.9 & 0.707 & 0.977 \\
METRO \cite{lin2021end} & 6.5 & 6.3 & 0.731 & 0.984 \\
Tang \textit{et al.} \cite{tang2021towards} & 6.7 & 6.7 & 0.724 & 0.981 \\
Lin \textit{et al.} \cite{lin2021mesh} & 5.9 & 6.0 & 0.764 & 0.986 \\
MobRecon  \cite{chen2022mobrecon} & 5.7 & 5.8 & 0.784 & 0.986 \\
AMVUR \cite{jiang2023probabilistic} & 6.2 & 6.1 & 0.767 & 0.987 \\
HaMeR \cite{pavlakos2024reconstructing} & 6.0 & 5.7 & 0.785 & 0.990 \\
Hamba\textsuperscript{†} \cite{dong2024hamba} & 5.8 & 5.5 & 0.798 & 0.991 \\
WiLoR\textsuperscript{†} \cite{potamias2024wilor} & 5.5 & \textbf{5.1} & \textbf{0.825} & \textbf{0.993} \\
HandOS\textsuperscript{†} \cite{chen2025handos} & \textbf{5.0} & 5.3 & 0.812 & 0.991 \\
\midrule
HHMR \cite{li2024hhmr} & 5.8 & 5.8 & - & - \\
HHMR (best) \cite{li2024hhmr} & 5.3 & 5.4 & - & - \\
\textbf{Proposed} & \textbf{5.0} & \underline{5.2} & \underline{0.816} & \underline{0.992} \\
\textbf{Proposed (best)} & 4.2 & 4.5 & 0.866 & 0.995  \\
\bottomrule
\end{tabular}
\vspace{2pt}
\caption{\textbf{Comparison with SOTAs on FreiHAND dataset.} We evaluate on the standard protocol and report metrics for predicted 3D joint and 3D mesh on FreiHAND. \textsuperscript{†} stands for using additional datasets. Methods above the line are deterministic methods, the others are probabilistic methods. "best" denotes an oracle best-of-32 selection.}
\label{tab:freihand_quant}
\vspace{-1cm}
\end{table}

\begin{table}[h]
\small
\centering
\setlength{\tabcolsep}{0.5pt}
\begin{tabular}{@{}l@{\hspace{3pt}}c@{\hspace{3pt}}c@{\hspace{3pt}}c@{\hspace{3pt}}c@{}}
\toprule
Method & P-MPJPE ↓ & P-MPVPE ↓ & F@5 ↑ & F@15 ↑  \\
\midrule
Hasson \textit{et al.} \cite{hasson2019learning} & 11.0 & 11.2 & 0.464 & 0.939  \\
Hampali \textit{et al.} \cite{hampali2020honnotate} & 10.7 & 10.6 & 0.506 & 0.942  \\
I2L-MeshNet \cite{moon2020i2l} & 11.2 & 13.0 & 0.409 & 0.932  \\
Pose2Mesh \cite{choi2020pose2mesh} & 12.5 & 12.7 & 0.441 & 0.909 \\
Liu \textit{et al.} \cite{liu2021semi} & 9.9 & 9.5 & 0.528 & 0.956  \\
I2UV-HandNet \cite{chen2021i2uv} & 9.9 & 10.1 & 0.500 & 0.943 \\
METRO \cite{lin2021end} & 10.4 & 11.1 & 0.484 & 0.946  \\
ArtiBoost \cite{yang2022artiboost} & 11.4 & 10.9 & 0.488 & 0.944  \\
MobRecon \cite{chen2022mobrecon} & 9.2 & 9.4 & 0.538 & 0.957 \\
Keypoint Trans. \cite{hampali2022keypoint} & 10.8 & - & - & - \\
HandOccNet \cite{park2022handoccnet} & 9.1 & 8.8 & 0.564 & 0.963  \\
AMVUR \cite{jiang2023probabilistic} & 8.3 & 8.2 & 0.608 & 0.965  \\
HaMeR \cite{pavlakos2024reconstructing} & 7.7 & 7.9 & 0.635 & 0.980  \\
Hamba\textsuperscript{†} \cite{dong2024hamba} & \textbf{7.5} & \underline{7.7} & \textbf{0.648} & \underline{0.982}  \\
WiLoR\textsuperscript{†} \cite{potamias2024wilor} & \textbf{7.5} & \underline{7.7} & \underline{0.646} & \textbf{0.983} \\
\midrule
\textbf{Proposed} & \textbf{7.5} & \textbf{7.5} & 0.633 & \underline{0.982} \\
\textbf{Proposed (best)} & 7.4 & 7.4 & 0.639 & 0.982 \\
\bottomrule
\end{tabular}
\vspace{2pt}
\caption{\textbf{Comparison with the state-of-the-art on the HO3Dv2 dataset.} We evaluate on the standard protocol and report metrics for predicted 3D joint and 3D mesh on HO3Dv2. \textsuperscript{†} stands for using additional datasets. "best" denotes an oracle best-of-32 selection.}
\label{tab:ho3d_quant}
\vspace{-0.5cm}
\end{table}

\subsection{Hand Pose Estimation} \label{sec:hpe}
We evaluate our cascaded diffusion framework on FreiHAND~\cite{zimmermann2019freihand}, HO3Dv2~\cite{hampali2020honnotate}, and DexYCB~\cite{chao2021dexycb} through both quantitative and qualitative experiments. As described in \cref{sec:method},
our inference pipeline generates multiple hypotheses at each stage: the joint diffusion model samples 50 joint hypotheses that are averaged into a stable estimate, and the Mesh LDM takes the aggregated joint and the input image to generate 50 latent vectors, whose average is decoded into the final hand mesh.

\vspace{2pt}
\noindent\textbf{Quantitative Results.} 
Tab.~\ref{tab:freihand_quant},~\ref{tab:ho3d_quant}, and~\ref{tab:dexycb} report the quantitative results on the three datasets. Our method achieves competitive performance across all three benchmarks, with the best results on DexYCB. Notably, recent methods such as Hamba~\cite{dong2024hamba}, WiLoR~\cite{potamias2024wilor}, and HandOS~\cite{chen2025handos} are trained on multiple datasets, whereas our model is trained only on the target dataset, indicating strong generalization under limited-data conditions. On the hand-object datasets, HO3Dv2 and DexYCB, where object-induced occlusion is common, our method remains competitive with the best approaches.

\vspace{2pt}
\noindent\textbf{Qualitative Results.}
\cref{fig:main_qualitative} presents qualitative comparisons on the three datasets. While all methods produce visually plausible meshes when rendered onto the image, the baselines tend to yield less natural finger articulation when viewed from the
side. For example, in the second row, where the ground truth is a \emph{rock} gesture, both HaMeR and WiLoR show slightly awkward finger bending that deviates from natural hand priors, whereas our method produces more realistic articulation by leveraging learned pose priors. Similar improvements are observed in other challenging examples, highlighting the advantage of our cascaded framework in producing more natural hand poses.

\begin{table}[t]
  \centering
  \begin{tabular}{l ccc}
    \toprule
    Method & P-MPJPE~$\downarrow$ & P-MPVPE~$\downarrow$ & AUC$_J$~$\uparrow$ \\
    \midrule
    Spurr \etal~\cite{spurr2020weakly}     & 6.8 & --  & 0.864 \\
    MobRecon~\cite{chen2022mobrecon}       & 6.4 & 5.6 & --    \\
    HandOccNet~\cite{park2022handoccnet}   & 5.8 & 5.5 & --    \\
    H2ONet~\cite{Xu_2023_CVPR}             & 5.7 & 5.5 & --    \\
    Zhou \etal~\cite{zhou2024simple}       & 5.5 & 5.5 & --    \\
    WiLoR~\cite{potamias2024wilor}         & \underline{4.9} & \underline{4.8} & \underline{0.902} \\
    HandOS~\cite{chen2025handos}           & 5.2 & 5.0 & 0.896 \\
    \midrule
    Proposed                               & \textbf{4.5} & \textbf{4.5} & \textbf{0.911} \\
    Proposed (best)                        & 4.2 & 4.2 & 0.917 \\
    \bottomrule
  \end{tabular}
  \caption{\textbf{Comparison with the state-of-the-art on the DexYCB dataset.} "best" denotes an oracle best-of-32 selection.}
  \label{tab:dexycb}
  \vspace{-0.5cm}
\end{table}


\vspace{2pt}
\noindent\textbf{Best-of-\textit{N} evaluation.}
We further examine whether the diverse joint hypotheses contain high-quality candidates through a best-of-$N$ evaluation. We sample $N$ joint hypotheses and reconstruct a mesh from each without averaging. We then report the accuracy of the mesh closest to the ground truth. In the main comparisons (Tab.~\ref{tab:freihand_quant},~\ref{tab:ho3d_quant}, and~\ref{tab:dexycb}), the "Proposed (best)" rows report the best of 32 hypotheses selected against ground truth, and are therefore oracle upper bounds. Under this setting, our method surpasses HHMR~\cite{li2024hhmr}, a probabilistic method that diffuses directly in 3D space, on FreiHAND (\cref{tab:freihand_quant}), where both methods select the best among 32 samples. This indicates that our cascaded latent design refines diverse hypotheses more effectively than single-stage 3D diffusion. The best results nonetheless verify a premise of our design: the hypotheses from the joint stage are not arbitrary noise but spread meaningfully around plausible poses, containing candidates close to the ground truth.
\begin{figure}[htb]
  \vspace{-0.2cm}
  \centering
   \includegraphics[width=0.9\linewidth]{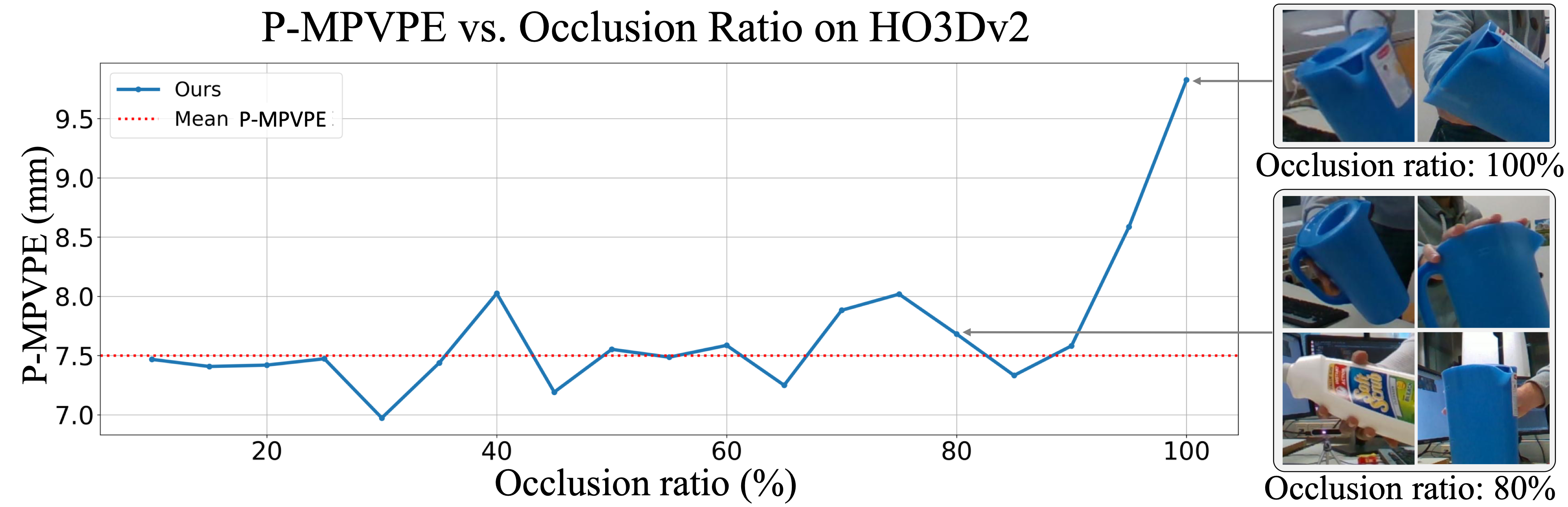}  
   \caption{\textbf{Robustness to occlusion on HO3Dv2.} P-MPVPE as a function of the occlusion ratio, defined as the fraction of the hand region overlapped by the manipulated object. Accuracy remains stable up to roughly $80\%$ occlusion and degrades only under extreme occlusion.
   }
   \label{fig:occ}
   \vspace{-0.2cm}
\end{figure}

\subsection{Robustness to Occlusion}
\label{sec:occlusion}
Hand-object interaction introduces object-induced occlusion, which makes both 2D-to-3D joint lifting and mesh reconstruction substantially harder. To examine how our framework behaves as occlusion becomes more severe, we evaluate on HO3Dv2~\cite{hampali2020honnotate} while varying the occlusion ratio, defined as the fraction of the hand region overlapped by the manipulated object. We group the test samples by their occlusion ratio and report P-MPVPE for each group.

As shown in \cref{fig:occ}, our method maintains stable accuracy up to roughly $80\%$ occlusion, with a sharper drop only under extreme occlusion where little visual evidence remains. We attribute this stability to our training scheme. The Mesh LDM is trained on diverse joint samples drawn from the joint diffusion model, so it learns to reconstruct meshes across a range of plausible joint conditions rather than from a single fixed input. When occlusion makes the coarse joint estimate less reliable, the mesh stage therefore still produces a plausible reconstruction. This robustness follows directly from our core design of using joint diversity as a training signal, rather than being an incidental property.

\subsection{Ablation study}
To analyze the contribution of each design choice, we conduct ablations on four aspects: (1) the joint diffusion model, (2) the number of sampled joint hypotheses, (3) the source of joint conditions, and (4) the representation used for mesh diffusion model.

\begin{table}[t]
    \centering
    \renewcommand{\arraystretch}{1.}
    \setlength{\tabcolsep}{6pt}
    \begin{tabular}{ccccccc}
        \toprule
        \multirow{2}{*}{\# samples}
        & \multicolumn{2}{c}{FreiHAND} & \multicolumn{2}{c}{HO3Dv2} & \multicolumn{2}{c}{DexYCB} \\
        \cmidrule(lr){2-3} \cmidrule(lr){4-5} \cmidrule(lr){6-7}
        & Average & Best & Average & Best & Average & Best \\
        \midrule
        1   & 5.04 & 5.04 & 7.90 & 7.90 & 7.61 & 7.61 \\
        10  & 5.01 & 4.78 & 7.89 & 7.80 & 7.60 & 7.40 \\
        50  & 5.01 & 4.67 & 7.89 & 7.75 & 7.59 & 7.30 \\
        100 & 5.01 & 4.63 & 7.89 & 7.74 & 7.59 & 7.26 \\
        200 & 5.01 & 4.59 & 7.89 & 7.72 & 7.59 & 7.23 \\
        \bottomrule
    \end{tabular}
    \vspace{2pt}
    \caption{\textbf{Effect of the number of samples from the joint diffusion model.} P-MPJPE (mm, ↓) of the Stage-1 joint output on FreiHAND, HO3Dv2 and DexYCB. "Best" is the oracle sample closest to ground truth.}    
    \label{tab:samples_joint}
    \vspace{-0.6cm}
\end{table}

\noindent \textbf{Performance of the joint diffusion model.} We assess the joint diffusion model by measuring P-MPJPE on its generated joints, using both the averaged and the best hypothesis (\cref{tab:samples_joint}). The model is competitive in direct pose estimation on FreiHAND, while errors are higher on the hand-object datasets HO3Dv2 and DexYCB, where object-induced occlusion and complex interaction make lifting 2D keypoints to 3D harder. Across the datasets, the averaged error stays nearly constant as the number of samples grows, whereas the best hypothesis steadily improves. This indicates that the samples are spread around the target rather than collapsing onto a single prediction. Since the quality of these joints directly affects the final mesh, this underlines the importance of reliable joint predictions in our cascaded framework.

\noindent \textbf{Number of sampling.} We analyze how the number of sampled joint hypotheses $N$ affects performance. As shown in \cref{tab:samples_joint}, increasing $N$ consistently improves the best-case P-MPJPE, while the averaged performance stays nearly unchanged. Larger $N$ therefore does not improve a single prediction better, instead expands the set of hypotheses and raises the chance that a high-quality one is present. We also observe that, for a fixed joint condition, the Mesh LDM itself produces little variance, since it primarily refines the given coarse joint into a mesh. These two observations together clarify the role of diversity in our framework: the diversity comes from the joint stage and is exploited as a signal during training, while the mesh stage acts as a stable refiner rather than a source of output variance. This is consistent with our inference scheme, where sampled joints and latents are aggregated into a stable estimate.

\begin{table}[h]
    \small
    \centering
    \begin{tabular}{@{}l@{\hspace{3pt}}c@{\hspace{3pt}}c@{\hspace{3pt}}c@{\hspace{3pt}}c@{}}
        \toprule
        \textbf{Joint condition}  & P-MPJPE ↓ & P-MPVPE ↓ & F@5 ↑ & F@15 ↑ \\
        \midrule
        off-the-shelf \cite{potamias2024wilor} & 5.1 & 5.4 & 0.800 & 0.991  \\
        Averaged joint  & 5.1 & 5.4 & 0.805 & 0.992  \\
        \midrule
        \textbf{Proposed} & 5.0 & 5.2 & 0.816 & 0.992 \\
        \textbf{Proposed (best)}  & 4.2 & 4.5 & 0.866 & 0.995  \\
        \bottomrule
    \end{tabular}
    \caption{\textbf{Ablation study of different joint condition sources on FreiHAND.} Conditioning the Mesh LDM on diverse joint hypotheses during training outperforms single-conditioned variants, improving accuracy and generalization across joint conditions. "best" denotes an oracle best-of-32 selection.}
    \label{tab:ablation}   
    \vspace{-0.2cm}
\end{table}

\par
\noindent \textbf{Source of joint conditions.} This ablation isolates the central claim of our framework: that exposing the mesh stage to \emph{diverse} joint conditions during training, rather than a single deterministic one, is what drives the improvement. We compare three sources of joint conditions for the Mesh LDM: (1) a single joint from an off-the-shelf  estimator~\cite{potamias2024wilor}, (2) the averaged joint from our joint diffusion model, and (3) diverse joint hypotheses from our full cascaded pipeline. During training, (1) and (2) condition the Mesh LDM on a single joint per image, whereas (3) conditions it on a joint sampled from the diffusion distribution, so it varies across iterations. At inference, all three condition on a single joint (the estimator output for (1), and the averaged joint for (2) and (3)), so the inference-time input is comparable across settings.
 
As shown in \cref{tab:ablation}, the two single-condition variants (1) and (2) perform almost identically (5.1 P-MPJPE), even though one uses an external estimator and the other uses our own averaged joint. What separates them from our full model is not the \emph{source} of the joint, but whether the mesh stage saw
a \emph{distribution} of joints during training: only the diverse-conditioned model (3) improves (5.0 P-MPJPE and consistently higher F-scores). This isolates diversity in the training condition as the decisive factor. We attribute this to
the mesh stage learning a joint-mesh mapping over a range of plausible joints instead of memorizing a single joint-to-mesh correspondence. As a result, the mesh stage generalizes better across the joint estimates it encounters at test time. In other words, accurate mesh reconstruction depends not only on the mesh model itself, but also on the distributional quality of the joints it is trained on. This is the purpose of our cascaded design. The two-stage structure is what makes diverse conditioning possible, and it is this diversity that drives the gain.

\begin{table}[]
    \small
    \centering
    \setlength{\tabcolsep}{0.7pt}
    \begin{tabular}{lcccc}
    \toprule
    \multirow{2}{*}{\normalsize \shortstack{Diffusion \\ space}} & \multicolumn{2}{c}{FreiHAND} & \multicolumn{2}{c}{HO3Dv2} \\
    \cmidrule(lr){2-3} \cmidrule(lr){4-5}
     & P-MPJPE ↓ & P-MPVPE ↓ & P-MPJPE ↓ & P-MPVPE ↓ \\
    \midrule
    MANO     & 5.70 & 5.82 & 7.81 & 7.81 \\
    Ours (latent)  & 5.00  & 5.23 & 7.50 & 7.52 \\
    \bottomrule
    \end{tabular}
    \vspace{2pt}
    \caption{\textbf{Quantitative comparison of diffusion in MANO space and our learned latent space.} We evaluate both representations using the same cascaded diffusion framework on the FreiHAND and HO3Dv2 datasets.}
    \label{tab:manolatent}
    \vspace{-0.4cm}
\end{table}

\vspace{2pt}
\noindent\textbf{Representation for mesh diffusion model.} Finally, we compare our learned latent representation with MANO~\cite{romero2017embodied} parameters as the target space for diffusion. As shown in \cref{tab:manolatent}, diffusion in our latent space achieves lower errors on both benchmarks. Although MANO provides a strong shape prior, every reconstruction is confined to its parameter manifold, which limits fine-grained articulation. Our latent space is free of this constraint and preserves both surface geometry and articulation. We therefore diffuse in this learned latent rather than in raw 3D vertex space, which tends to lose surface and pose structure (\cref{fig:forward_process}(a)). Overall, a learned latent space is an effective target for diffusion-based mesh reconstruction in our pipeline.

\section{Conclusions}
We presented a coarse-to-fine cascaded diffusion framework for 3D hand pose estimation, combining a joint diffusion model and a Mesh Latent Diffusion Model (Mesh LDM). The joint diffusion model generates diverse 3D joint hypotheses, and the Mesh LDM reconstructs a 3D hand mesh conditioned on a joint sample from these hypotheses. Rather than treating the diverse hypotheses as final outputs, we use the hypotheses as a training signal for the mesh stage to learn distribution-aware joint-mesh relationships robust to variations in coarse predictions.

Extensive experiments on the FreiHAND, HO3Dv2, and DexYCB benchmarks show that our method achieves strong performance, reaching the best results on DexYCB, and remains stable under occlusion and pose ambiguity.

For future work, we plan to extend our framework to two-hand interaction and to explicit hand--object modeling, jointly reconstructing the hand and the manipulated object to better handle complex real-world scenarios.

\section*{Acknowledgements}
This work was supported by NST grant (CRC 21015, CRC23023-000, MSIT), IITP grant (RS-2023-00228996, RS-2024- 00459749, RS-2025-25443318, RS-2025-25441313, RS-2026-25526850, RS-2026-25522885, MSIT) , KOCCA grant (RS-2024-00442308, MCST) and InnoCORE program (N10260110, MSIT)


%
%
\bibliographystyle{splncs04}
\bibliography{main}

@String(IJCV = {Int. J. Comput. Vis.})

@String(CVPR= {IEEE Conf. Comput. Vis. Pattern Recog.})

@String(ICCV= {Int. Conf. Comput. Vis.})

@String(ECCV= {Eur. Conf. Comput. Vis.})

@String(NIPS= {Adv. Neural Inform. Process. Syst.})

@String(TOG= {ACM Trans. Graph.})

@String(ICIP = {IEEE Int. Conf. Image Process.})

@String(ICLR = {Int. Conf. Learn. Represent.})

@String(AAAI = {AAAI})

@String(VR   = {Vis. Res.})

@String(IJCV  = {IJCV})

@String(CVPR  = {CVPR})

@String(ICCV  = {ICCV})

@String(ECCV  = {ECCV})

@String(NIPS  = {NeurIPS})

@String(TOG   = {ACM TOG})

@String(ICIP  = {ICIP})

@String(ICLR  = {ICLR})

@inproceedings{moon2020i2l,
  title={I2l-meshnet: Image-to-lixel prediction network for accurate 3d human pose and mesh estimation from a single rgb image},
  author={Moon, Gyeongsik and Lee, Kyoung Mu},
  booktitle=ECCV,
  year=2020
}

@inproceedings{choi2020pose2mesh,
  title={Pose2mesh: Graph convolutional network for 3d human pose and mesh recovery from a 2d human pose},
  author={Choi, Hongsuk and Moon, Gyeongsik and Lee, Kyoung Mu},
  booktitle=ECCV,
  year=2020
}

@inproceedings{chen2021i2uv,
  title={I2uv-handnet: Image-to-uv prediction network for accurate and high-fidelity 3d hand mesh modeling},
  author={Chen, Ping and Chen, Yujin and Yang, Dong and Wu, Fangyin and Li, Qin and Xia, Qingpei and Tan, Yong},
  booktitle=ICCV,
  year=2021
}

@inproceedings{lin2021end,
  title={End-to-end human pose and mesh reconstruction with transformers},
  author={Lin, Kevin and Wang, Lijuan and Liu, Zicheng},
  booktitle=CVPR,
  year=2021
}

@inproceedings{tang2021towards,
  title={Towards accurate alignment in real-time 3d hand-mesh reconstruction},
  author={Tang, Xiao and Wang, Tianyu and Fu, Chi-Wing},
  booktitle=ICCV,
  year=2021
}

@inproceedings{lin2021mesh,
  title={Mesh graphormer},
  author={Lin, Kevin and Wang, Lijuan and Liu, Zicheng},
  booktitle=ICCV,
  year=2021
}

@inproceedings{chen2022mobrecon,
  title={Mobrecon: Mobile-friendly hand mesh reconstruction from monocular image},
  author={Chen, Xingyu and Liu, Yufeng and Dong, Yajiao and Zhang, Xiong and Ma, Chongyang and Xiong, Yanmin and Zhang, Yuan and Guo, Xiaoyan},
  booktitle=CVPR,
  year=2022
}

@inproceedings{jiang2023probabilistic,
  title={A probabilistic attention model with occlusion-aware texture regression for 3D hand reconstruction from a single RGB image},
  author={Jiang, Zheheng and Rahmani, Hossein and Black, Sue and Williams, Bryan M},
  booktitle=CVPR,
  year=2023
}

@inproceedings{pavlakos2024reconstructing,
  title={Reconstructing hands in 3d with transformers},
  author={Pavlakos, Georgios and Shan, Dandan and Radosavovic, Ilija and Kanazawa, Angjoo and Fouhey, David and Malik, Jitendra},
  booktitle=CVPR,
  year=2024
}

@inproceedings{potamias2024wilor,
  title={Wilor: End-to-end 3d hand localization and reconstruction in-the-wild},
  author={Potamias, Rolandos Alexandros and Zhang, Jinglei and Deng, Jiankang and Zafeiriou, Stefanos},
  booktitle=CVPR,
  year=2025
}

@article{dong2024hamba,
  title={Hamba: Single-view 3d hand reconstruction with graph-guided bi-scanning mamba},
  author={Dong, Haoye and Chharia, Aviral and Gou, Wenbo and Carrasco, Francisco Vicente and De la Torre, Fernando},
  journal=NIPS,
  year=2024
}

@inproceedings{li2024hhmr,
  title={HHMR: Holistic Hand Mesh Recovery by Enhancing the Multimodal Controllability of Graph Diffusion Models},
  author={Li, Mengcheng and Zhang, Hongwen and Zhang, Yuxiang and Shao, Ruizhi and Yu, Tao and Liu, Yebin},
  booktitle=CVPR,
  year=2024
}

@inproceedings{liu2021semi,
  title={Semi-supervised 3d hand-object poses estimation with interactions in time},
  author={Liu, Shaowei and Jiang, Hanwen and Xu, Jiarui and Liu, Sifei and Wang, Xiaolong},
  booktitle=CVPR,
  year=2021
}

@inproceedings{jain2024video,
  title={Video interpolation with diffusion models},
  author={Jain, Siddhant and Watson, Daniel and Tabellion, Eric and Poole, Ben and Kontkanen, Janne and others},
  booktitle=CVPR,
  year={2024}
}

@article{li2025likelihood,
  title={Likelihood training of cascaded diffusion models via hierarchical volume-preserving maps},
  author={Li, Henry and Basri, Ronen and Kluger, Yuval},
  journal=ICLR,
  year={2024}
}

@article{salimans2022progressive,
  title={Progressive distillation for fast sampling of diffusion models},
  author={Salimans, Tim and Ho, Jonathan},
  journal=ICLR,
  year={2022}
}

@inproceedings{shabani2024visual,
  title={Visual layout composer: Image-vector dual diffusion model for design layout generation},
  author={Shabani, Mohammad Amin and Wang, Zhaowen and Liu, Difan and Zhao, Nanxuan and Yang, Jimei and Furukawa, Yasutaka},
  booktitle=CVPR,
  year={2024}
}

@inproceedings{park2022handoccnet,
  title={Handoccnet: Occlusion-robust 3d hand mesh estimation network},
  author={Park, JoonKyu and Oh, Yeonguk and Moon, Gyeongsik and Choi, Hongsuk and Lee, Kyoung Mu},
  booktitle=CVPR,
  year=2022
}

@inproceedings{cheng2024handdiff,
  title={Handdiff: 3d hand pose estimation with diffusion on image-point cloud},
  author={Cheng, Wencan and Tang, Hao and Van Gool, Luc and Ko, Jong Hwan},
  booktitle=CVPR,
  year={2024}
}

@inproceedings{hasson2019learning,
  title={Learning joint reconstruction of hands and manipulated objects},
  author={Hasson, Yana and Varol, Gul and Tzionas, Dimitrios and Kalevatykh, Igor and Black, Michael J and Laptev, Ivan and Schmid, Cordelia},
  booktitle=CVPR,
  year=2019
}

@inproceedings{yang2022artiboost,
  title={Artiboost: Boosting articulated 3d hand-object pose estimation via online exploration and synthesis},
  author={Yang, Lixin and Li, Kailin and Zhan, Xinyu and Lv, Jun and Xu, Wenqiang and Li, Jiefeng and Lu, Cewu},
  booktitle=CVPR,
  year=2022
}

@inproceedings{hampali2022keypoint,
  title={Keypoint transformer: Solving joint identification in challenging hands and object interactions for accurate 3d pose estimation},
  author={Hampali, Shreyas and Sarkar, Sayan Deb and Rad, Mahdi and Lepetit, Vincent},
  booktitle=CVPR,
  year=2022
}

@inproceedings{zimmermann2019freihand,
  title={FreiHAND: A dataset for markerless capture of hand pose and shape from single rgb images},
  author={Zimmermann, Christian and Ceylan, Duygu and Yang, Jimei and Russell, Bryan and Argus, Max and Brox, Thomas},
  booktitle=ICCV,
  year=2019
}

@inproceedings{hampali2020honnotate,
  title={Honnotate: A method for 3d annotation of hand and object poses},
  author={Hampali, Shreyas and Rad, Mahdi and Oberweger, Markus and Lepetit, Vincent},
  booktitle=CVPR,
  year=2020
}

@article{ho2022cascaded,
  title={Cascaded diffusion models for high fidelity image generation},
  author={Ho, Jonathan and Saharia, Chitwan and Chan, William and Fleet, David J and Norouzi, Mohammad and Salimans, Tim},
  journal={JMLR},
  year={2022}
}

@article{zhang2023i2vgen,
  title={I2vgen-xl: High-quality image-to-video synthesis via cascaded diffusion models},
  author={Zhang, Shiwei and Wang, Jiayu and Zhang, Yingya and Zhao, Kang and Yuan, Hangjie and Qin, Zhiwu and Wang, Xiang and Zhao, Deli and Zhou, Jingren},
  journal={arXiv preprint arXiv:2311.04145},
  year={2023}
}

@inproceedings{lee2024interhandgen,
  title={Interhandgen: Two-hand interaction generation via cascaded reverse diffusion},
  author={Lee, Jihyun and Saito, Shunsuke and Nam, Giljoo and Sung, Minhyuk and Kim, Tae-Kyun},
  booktitle=CVPR,
  year={2024}
}

@inproceedings{koo2023salad,
  title={Salad: Part-level latent diffusion for 3d shape generation and manipulation},
  author={Koo, Juil and Yoo, Seungwoo and Nguyen, Minh Hieu and Sung, Minhyuk},
  booktitle=ICCV,
  year={2023}
}

@article{shafir2023human,
  title={Human motion diffusion as a generative prior},
  author={Shafir, Yonatan and Tevet, Guy and Kapon, Roy and Bermano, Amit H},
  journal=ICLR,
  year={2023}
}

@inproceedings{guleryuz2018fast,
  title={Fast lifting for 3d hand pose estimation in ar/vr applications},
  author={Guleryuz, Onur G and Kaeser-Chen, Christine},
  booktitle=ICIP,
  year={2018}
}

@inproceedings{shi2022grasping,
  title={Grasping 3D Objects With Virtual Hand in VR Environment},
  author={Shi, Yinghan and Zhao, Lizhi and Lu, Xuequan and Hoang, Thuong and Wang, Meili},
  booktitle={SIGGRAPH},
  year={2022}
}

@inproceedings{li2019vision,
  title={Vision-based teleoperation of shadow dexterous hand using end-to-end deep neural network},
  author={Li, Shuang and Ma, Xiaojian and Liang, Hongzhuo and G{\"o}rner, Michael and Ruppel, Philipp and Fang, Bin and Sun, Fuchun and Zhang, Jianwei},
  booktitle=ICRA,
  year={2019}
}

@inproceedings{handa2020dexpilot,
  title={Dexpilot: Vision-based teleoperation of dexterous robotic hand-arm system},
  author={Handa, Ankur and Van Wyk, Karl and Yang, Wei and Liang, Jacky and Chao, Yu-Wei and Wan, Qian and Birchfield, Stan and Ratliff, Nathan and Fox, Dieter},
  booktitle=ICRA,
  year={2020}
}

@article{lopez2023dexterous,
  title={Dexterous Object Manipulation with an Anthropomorphic Robot Hand via Natural Hand Pose Transformer and Deep Reinforcement Learning},
  author={Lopez, Patricio Rivera and Oh, Ji-Heon and Jeong, Jin Gyun and Jung, Hwanseok and Lee, Jin Hyuk and Jaramillo, Ismael Espinoza and Chola, Channabasava and Lee, Won Hee and Kim, Tae-Seong},
  journal={Applied Sciences},
  year={2023}
}

@inproceedings{sharma2019grasping,
  title={Grasping microgestures: Eliciting single-hand microgestures for handheld objects},
  author={Sharma, Adwait and Roo, Joan Sol and Steimle, J{\"u}rgen},
  booktitle={Proceedings of the 2019 CHI Conference on Human Factors in Computing Systems},
  year={2019}
}

@inproceedings{sharma2021solofinger,
  title={SoloFinger: Robust microgestures while grasping everyday objects},
  author={Sharma, Adwait and Hedderich, Michael A and Bhardwaj, Divyanshu and Fruchard, Bruno and McIntosh, Jess and Nittala, Aditya Shekhar and Klakow, Dietrich and Ashbrook, Daniel and Steimle, J{\"u}rgen},
  booktitle={Proceedings of the 2021 CHI Conference on Human Factors in Computing Systems},
  year={2021}
}

@article{ho2020denoising,
  title={Denoising diffusion probabilistic models},
  author={Ho, Jonathan and Jain, Ajay and Abbeel, Pieter},
  journal=NIPS,
  year={2020}
}

@inproceedings{rombach2022high,
  title={High-resolution image synthesis with latent diffusion models},
  author={Rombach, Robin and Blattmann, Andreas and Lorenz, Dominik and Esser, Patrick and Ommer, Bj{\"o}rn},
  booktitle=CVPR,
  year={2022}
}

@inproceedings{
tevet2023human,
title={Human Motion Diffusion Model},
author={Guy Tevet and Sigal Raab and Brian Gordon and Yoni Shafir and Daniel Cohen-or and Amit Haim Bermano},
booktitle=ICLR,
year={2023}
}

@article{liang2024intergen,
  title={Intergen: Diffusion-based multi-human motion generation under complex interactions},
  author={Liang, Han and Zhang, Wenqian and Li, Wenxuan and Yu, Jingyi and Xu, Lan},
  journal=IJCV,
  year={2024}
}

@inproceedings{shan2023diffusion,
  title={Diffusion-based 3d human pose estimation with multi-hypothesis aggregation},
  author={Shan, Wenkang and Liu, Zhenhua and Zhang, Xinfeng and Wang, Zhao and Han, Kai and Wang, Shanshe and Ma, Siwei and Gao, Wen},
  booktitle=ICCV,
  year={2023}
}

@inproceedings{gong2023diffpose,
  title={Diffpose: Toward more reliable 3d pose estimation},
  author={Gong, Jia and Foo, Lin Geng and Fan, Zhipeng and Ke, Qiuhong and Rahmani, Hossein and Liu, Jun},
  booktitle=CVPR,
  year={2023}
}

@inproceedings{feng2023diffpose,
  title={Diffpose: Spatiotemporal diffusion model for video-based human pose estimation},
  author={Feng, Runyang and Gao, Yixing and Tse, Tze Ho Elden and Ma, Xueqing and Chang, Hyung Jin},
  booktitle=ICCV,
  year={2023}
}

@inproceedings{holmquist2023diffpose,
  title={Diffpose: Multi-hypothesis human pose estimation using diffusion models},
  author={Holmquist, Karl and Wandt, Bastian},
  booktitle=ICCV,
  year={2023}
}

@inproceedings{ivashechkin2023denoising,
  title={Denoising Diffusion for 3D Hand Pose Estimation from Images},
  author={Ivashechkin, Maksym and Mendez, Oscar and Bowden, Richard},
  booktitle=ICCV,
  year={2023}
}

@inproceedings{ramesh2021zero,
  title={Zero-shot text-to-image generation},
  author={Ramesh, Aditya and Pavlov, Mikhail and Goh, Gabriel and Gray, Scott and Voss, Chelsea and Radford, Alec and Chen, Mark and Sutskever, Ilya},
  booktitle=ICML,
  year={2021}
}

@inproceedings{cho2023generative,
  title={Generative approach for probabilistic human mesh recovery using diffusion models},
  author={Cho, Hanbyel and Kim, Junmo},
  booktitle={ICCVW},
  year={2023}
}

@article{dposer,
  title={DPoser: Diffusion Model as Robust 3D Human Pose Prior},
  author={Lu, Junzhe and Lin, Jing and Dou, Hongkun and Zeng, Ailing and Deng, Yue and Zhang, Yulun and Wang, Haoqian},
  journal={arxiv:2312.05541},
  year={2023}
}

@inproceedings{foo2023distribution,
  title={Distribution-aligned diffusion for human mesh recovery},
  author={Foo, Lin Geng and Gong, Jia and Rahmani, Hossein and Liu, Jun},
  booktitle=ICCV,
  year={2023}
}

@inproceedings{gong2019spiralnet,
  title={Spiralnet++: A fast and highly efficient mesh convolution operator},
  author={Gong, Shunwang and Chen, Lei and Bronstein, Michael and Zafeiriou, Stefanos},
  booktitle=ICCV,
  year={2019}
}

@inproceedings{peebles2023scalable,
  title={Scalable diffusion models with transformers},
  author={Peebles, William and Xie, Saining},
  booktitle=ICCV,
  year={2023}
}

@article{song2020denoising,
  title={Denoising diffusion implicit models},
  author={Song, Jiaming and Meng, Chenlin and Ermon, Stefano},
  journal=ICLR,
  year={2020}
}

@inproceedings{zhang2022mixste,
  title={Mixste: Seq2seq mixed spatio-temporal encoder for 3d human pose estimation in video},
  author={Zhang, Jinlu and Tu, Zhigang and Yang, Jianyu and Chen, Yujin and Yuan, Junsong},
  booktitle=CVPR,
  year={2022}
}

@inproceedings{ge20193d,
  title={3d hand shape and pose estimation from a single rgb image},
  author={Ge, Liuhao and Ren, Zhou and Li, Yuncheng and Xue, Zehao and Wang, Yingying and Cai, Jianfei and Yuan, Junsong},
  booktitle=CVPR,
  year={2019}
}

@inproceedings{kulon2020weakly,
  title={Weakly-supervised mesh-convolutional hand reconstruction in the wild},
  author={Kulon, Dominik and Guler, Riza Alp and Kokkinos, Iasonas and Bronstein, Michael M and Zafeiriou, Stefanos},
  booktitle=CVPR,
  year={2020}
}

@inproceedings{baek2019pushing,
  title={Pushing the envelope for rgb-based dense 3d hand pose estimation via neural rendering},
  author={Baek, Seungryul and Kim, Kwang In and Kim, Tae-Kyun},
  booktitle=CVPR,
  year={2019}
}

@inproceedings{boukhayma20193d,
  title={3d hand shape and pose from images in the wild},
  author={Boukhayma, Adnane and Bem, Rodrigo de and Torr, Philip HS},
  booktitle=CVPR,
  year={2019}
}

@inproceedings{zhang2019end,
  title={End-to-end hand mesh recovery from a monocular rgb image},
  author={Zhang, Xiong and Li, Qiang and Mo, Hong and Zhang, Wenbo and Zheng, Wen},
  booktitle=ICCV,
  year={2019}
}

@inproceedings{perez2018film,
  title={Film: Visual reasoning with a general conditioning layer},
  author={Perez, Ethan and Strub, Florian and De Vries, Harm and Dumoulin, Vincent and Courville, Aaron},
  booktitle=AAAI,
  year={2018}
}

@article{xu2022vitpose,
  title={Vitpose: Simple vision transformer baselines for human pose estimation},
  author={Xu, Yufei and Zhang, Jing and Zhang, Qiming and Tao, Dacheng},
  journal=NIPS,
  year={2022}
}

@inproceedings{he2016deep,
  title={Deep residual learning for image recognition},
  author={He, Kaiming and Zhang, Xiangyu and Ren, Shaoqing and Sun, Jian},
  booktitle=CVPR,
  year={2016}
}

@article{loshchilov2017decoupled,
  title={Decoupled weight decay regularization},
  author={Loshchilov, Ilya and Hutter, Frank},
  journal={arXiv preprint arXiv:1711.05101},
  year={2017}
}

@article{romero2017embodied,
  title={Embodied hands: modeling and capturing hands and bodies together},
  author={Romero, Javier and Tzionas, Dimitrios and Black, Michael J},
  journal={ACM Transactions on Graphics (TOG)},
  year={2017}
}

@inproceedings{chen2025handos,
  title={HandOS: 3D Hand Reconstruction in One Stage},
  author={Chen, Xingyu and Song, Zhuheng and Jiang, Xiaoke and Hu, Yaoqing and Yu, Junzhi and Zhang, Lei},
  booktitle=CVPR,
  year={2025}
}

@inproceedings{lee2025rewind,
  title={REWIND: Real-Time Egocentric Whole-Body Motion Diffusion with Exemplar-Based Identity Conditioning},
  author={Lee, Jihyun and Xu, Weipeng and Richard, Alexander and Wei, Shih-En and Saito, Shunsuke and Bai, Shaojie and Wang, Te-Li and Sung, Minhyuk and Kim, Tae-Kyun and Saragih, Jason},
  booktitle=CVPR,
  year={2025}
}

@inproceedings{cho2024dense,
  title={Dense hand-object (ho) graspnet with full grasping taxonomy and dynamics},
  author={Cho, Woojin and Lee, Jihyun and Yi, Minjae and Kim, Minje and Woo, Taeyun and Kim, Donghwan and Ha, Taewook and Lee, Hyokeun and Ryu, Je-Hwan and Woo, Woontack and others},
  booktitle=ECCV,
  year={2024}
}

@article{lee2023fourierhandflow,
  title={Fourierhandflow: Neural 4d hand representation using fourier query flow},
  author={Lee, Jihyun and Jang, Junbong and Kim, Donghwan and Sung, Minhyuk and Kim, Tae-Kyun},
  journal=NIPS,
  year={2023}
}

@inproceedings{lee2023im2hands,
  title={Im2hands: Learning attentive implicit representation of interacting two-hand shapes},
  author={Lee, Jihyun and Sung, Minhyuk and Choi, Honggyu and Kim, Tae-Kyun},
  booktitle=CVPR,
  year={2023}
}

@inproceedings{baek2020weakly,
  title={Weakly-supervised domain adaptation via gan and mesh model for estimating 3d hand poses interacting objects},
  author={Baek, Seungryul and Kim, Kwang In and Kim, Tae-Kyun},
  booktitle=CVPR,
  year={2020}
}

@inproceedings{armagan2020measuring,
  title={Measuring generalisation to unseen viewpoints, articulations, shapes and objects for 3D hand pose estimation under hand-object interaction},
  author={Armagan, Anil and Garcia-Hernando, Guillermo and Baek, Seungryul and Hampali, Shreyas and Rad, Mahdi and Zhang, Zhaohui and Xie, Shipeng and Chen, MingXiu and Zhang, Boshen and Xiong, Fu and others},
  booktitle=ECCV,
  year={2020}
}

@inproceedings{chao2021dexycb,
  title={Dexycb: A benchmark for capturing hand grasping of objects},
  author={Chao, Yu-Wei and Yang, Wei and Xiang, Yu and Molchanov, Pavlo and Handa, Ankur and Tremblay, Jonathan and Narang, Yashraj S and Van Wyk, Karl and Iqbal, Umar and Birchfield, Stan and others},
  booktitle=CVPR,
  year={2021}
}

@inproceedings{zhou2024simple,
  title={A simple baseline for efficient hand mesh reconstruction},
  author={Zhou, Zhishan and Zhou, Shihao and Lv, Zhi and Zou, Minqiang and Tang, Yao and Liang, Jiajun},
  booktitle=CVPR,
  year={2024}
}

@inproceedings{Xu_2023_CVPR,
    author    = {Xu, Hao and Wang, Tianyu and Tang, Xiao and Fu, Chi-Wing},
    title     = {H2ONet: Hand-Occlusion-and-Orientation-Aware Network for Real-Time 3D Hand Mesh Reconstruction},
    booktitle = CVPR,
    year      = {2023},
}

@inproceedings{spurr2020weakly,
  title={Weakly supervised 3d hand pose estimation via biomechanical constraints},
  author={Spurr, Adrian and Iqbal, Umar and Molchanov, Pavlo and Hilliges, Otmar and Kautz, Jan},
  booktitle=ECCV,
  year={2020},
}

@inproceedings{woo2024hand,
  title={Hand-object reconstruction via interaction-aware graph attention mechanism},
  author={Woo, Taeyun and Kim, Tae-Kyun and Park, Jinah},
  booktitle=ICIP,
  year={2024},
  organization={IEEE}
}


\clearpage
\setcounter{page}{1}
\setcounter{section}{0}
\maketitlesupplementary
\renewcommand{\thesection}{\Alph{section}}

In this supplementary material, we provide implementation details of our cascaded diffusion model. We also present additional results on the FreiHAND~\cite{zimmermann2019freihand} and HO3Dv2~\cite{hampali2020honnotate} datasets.

\section{Implementation details} \label{sec:id}
\subsection{Joint diffusion model}
Our joint diffusion model is adapted from the D3DP framework~\cite{shan2023diffusion}, originally designed for lifting 2D human pose sequences to 3D using the MixSTE backbone~\cite{zhang2022mixste}. However, since our goal is estimating single-frame 3D hand poses, we modify the sequence length to 1. The model employs a hidden dimension of 512 and includes 8 MixSTE blocks. We normalize 3D hand joints during training.

The model is trained using only the diffusion loss $\mathcal{L}_{\text{DDPM}}$, with a linear noise scheduler ($\beta \in [0.0001, 0.01]$) with 1000 diffusion timesteps. Input 2D keypoints are obtained from an off-the-shelf estimator~\cite{potamias2024wilor}. To enhance generalization, we apply data augmentation by randomly rotating the 2D keypoints and 3D hand joints within [-60$^{\circ}$, 60$^{\circ}$] and scaling the 2D keypoints within [0.9, 1.1]. The joint diffusion model is trained for 250K steps with an initial learning rate of 1e-4, which decays by a factor of 0.9 every 20K steps using a step-based learning rate schedule.

\subsection{Mesh AutoEncoder}
The Mesh AutoEncoder (Mesh AE) is based on SpiralNet++ framework~\cite{gong2019spiralnet}, which encodes 3D mesh using spiral convolutions. Given a hand mesh with vertex positions $V \in \mathbb{R}^{778 \times 3}$, the encoder $\mathcal{E}$ compresses the mesh into a latent representation $x \in \mathbb{R}^{168}$, while the decoder $\mathcal{D}$ reconstructs the hand mesh from the latent vector:

\begin{equation}
    x=\mathcal{E}(V), \hat{V}=\mathcal{D}(x).
\end{equation}
Note that the hand mesh $V$ is mean-centered, \ie $\bar{V}=\mathbf{0}$.
We employ the following loss terms during training:
\begin{itemize}
    \item \textbf{Vertex Loss $\mathcal{L}_V$}: L1 loss between the ground-truth mesh vertices $V$ and the reconstructed vertices $\hat{V}$, encouraging accurate mesh reconstruction.
    \item \textbf{Joint Loss $\mathcal{L}_J$}: L1 loss between the ground-truth hand joint $J$ and the reconstructed joint $\hat{J}=\mathcal{J}\hat{V}$. $\mathcal{J}$ is a joint regressor matrix.
    \item \textbf{KL Regularization $\mathcal{L}_{KL}$}: We apply KL divergence to the latent vector of AE to follow a Gaussian distribution, which improves the generalization for Mesh LDM.
    \item \textbf{Loss configuration}: \begin{equation}
        \mathcal{L} = \lambda_{V}\mathcal{L}_{V} + \lambda_{J}\mathcal{L}_{J}+\lambda_{KL}\mathcal{L}_{KL}
    \end{equation} where $\lambda_{V}=1, \lambda_{J}=0.5, \lambda_{KL}=1e-3 $.
\end{itemize}

Mesh AE is trained for 1000 epochs with batch size of 50 with AdamW optimizer~\cite{loshchilov2017decoupled}. The initial learning rate is 1e-3 and decays a factor of 0.9 every 50 epochs. Note that for HO3Dv2~\cite{hampali2020honnotate} training, the initial learning rate is 1e-4. 

\subsection{Mesh LDM}
Mesh LDM reconstructs the latent vector of a 3D hand mesh by denoising a noisy latent vector: $x_t=\sqrt{\bar{\alpha}_t} x_0 + \sqrt{1-\bar{\alpha}_t}\epsilon$, where $\bar{\alpha}_t=\prod_{i=1}^{t} \alpha_i$,$\epsilon\sim N(0, I)$, and $\alpha_t$ is noise variance schedule at timestep $t$. It is conditioned on: (1) the reconstructed 3D joint $\hat{J}_0$ from the joint diffusion model and (2) image features $\mathcal{I}$ extracted from an image encoder. The image features $\mathcal{I}$ consist of four levels of extracted features: $\mathcal{I}=\{\mathcal{I}_1, \mathcal{I}_2, \mathcal{I}_3, \mathcal{I}_4\}$. 

During training on FreiHAND~\cite{zimmermann2019freihand}, we use ViT-based features from an off-the-shelf encoder~\cite{potamias2024wilor}. For HO3Dv2~\cite{hampali2020honnotate}, we use ResNet-50~\cite{he2016deep}, where each stage’s output is treated as a feature level. In the ViT-based case, a single global image feature is upsampled using three deconvolution layers. 

\noindent \textbf{Architecture.} Mesh LDM follows the DiT framework~\cite{peebles2023scalable}, employing a transformer-based architecture. The input latent vector is repeated 21 times, and concatenated with the denoised joint $\hat{J}_0$. The resulting input tensor has a shape of $\mathbb{R}^{171 \times  21}$, where 171 is the channel dimension, and 21 is the sequence length. The input tensor is tokenized, and the channel dimension is expanded to 512. The Mesh LDM consists of 12 transformer blocks with 16 attention heads. Each Mesh LDM block processes the input as follows:
\begin{enumerate}
    \item[(1)] \textbf{Cross-attention}: conducting cross-attention with each image feature and concatenate the results.
    \item[(2)] \textbf{Self-attention and MLP}: Similar to DiT, these layers refine the latent representation.
    \item[(3)] \textbf{Output layer}: The denoised latent $\hat{x}_0$ outputs through reshape function. 
\end{enumerate}
We apply Adaptive layer normalization~\cite{perez2018film} between layers to each level of layers. Finally, the output dimension is 8, and the final tensor is flatten the final tensor to reconstruct 168-dimensional latent vector.

\noindent \textbf{Training.} The Mesh LDM is trained for 100K steps with a learning rate of 1e-4, using 1000 diffusion timesteps, and decays by a factor of 0.9 every 5K steps. The training loss includes:
\begin{itemize}
    \item \textbf{Diffusion loss $\mathcal{L}_{DDPM}$}: L2 loss between the ground-truth latent vector $x_0$ and the reconstructed latent $\hat{x}_0$.
    \item \textbf{Vertex Loss $\mathcal{L}_V$}: L1 loss between the ground-truth mesh vertices $V$ and the reconstructed vertices $\hat{V}$, encouraging accurate mesh reconstruction.
    \item \textbf{Joint Loss $\mathcal{L}_J$}: L1 loss between the ground-truth hand joint $J$ and the reconstructed joint $\hat{J}=\mathcal{J}\hat{V}$. $\mathcal{J}$ is a joint regressor matrix.
    \item \textbf{Loss configuration}: \begin{equation}
        \mathcal{L} = \lambda_{DDPM}\mathcal{L}_{DDPM}+\lambda_{V}\mathcal{L}_{V} + \lambda_{J}\mathcal{L}_{J},
    \end{equation} where $\lambda_{DDPM}=1, \lambda_{V}=10, \lambda_{J}=5$.
\end{itemize}
We also apply rotation augmentations to images and corresponding 3D hand joints and hand mesh.

\subsection{Details for MANO Mesh LDM}
For the ablation study, we also evaluate a variant of Mesh LDM that predicts latent vector in the MANO parameter~\cite{romero2017embodied} space. In implementation, the 58-dimensional MANO parameters are repeated 21 times, similar to original models. Then, the repeated vectors and the 3D joint coordinates (21$\times$3) are concatenated, forming a 61-dimensional latent vector with 21 sequence lengths ($\mathbb{R}^{61 \times 21}$). As the output vector's shape is $\mathbb{R}^{168}$, we change the shape of it with MLP layer to form $\mathbb{R}^{58}$ vectors.

\section{Additional Results} \label{sec:aqr}

\begin{table}[h]
    \centering
    \begin{tabular}{c|cccc}
        \toprule
        \shortstack{Image \\ encoders} & P-MPJPE ↓ &  P-MPVPE ↓ & F@5 ↑ & F@15 ↑ \\ 
        \hline
        off-the-shelf  &  5.00 & 5.23 & 0.816 & 0.992 \\  
        ViTPose-B  & 5.02  & 5.26 & 0.811 & 0.992 \\  
        \bottomrule
    \end{tabular}
    
    \caption{\textbf{Comparison of mesh reconstruction performance using different image encoders.} Both the off-the-shelf encoder and ViTPose-B yield comparable results across all metrics, demonstrating the robustness of our cascaded framework to encoder variation on FreiHAND dataset.}
    \label{tab:encoders}
\end{table}

\subsection{Variant of image encoder}
To assess the effect of the image encoder, we additionally train our model using a ViTPose-B encoder~\cite{xu2022vitpose}. Table~\ref{tab:encoders} reports the comparison results under the same evaluation protocol as the main experiments. Although ViTPose-B is trained on a smaller dataset than the off-the-shelf encoder, the performance differences are comparable, indicating that our cascaded framework is robust to variations in encoder architecture and generalizes well across different image encoders.

\begin{table}[h]
    \centering
    \renewcommand{\arraystretch}{1.2}
    \begin{tabular}{c|cccc}
        \toprule
        \shortstack{Number of \\ samples} & P-MPJPE ↓ &  P-MPVPE ↓ & F@5 ↑ & F@15 ↑ \\ 
        \hline
        1  &  5.0 & 5.2 & 0.816 & 0.992 \\  
        5  &  4.7 & 4.7 & 0.835 & 0.994 \\  
        10 &  4.6 & 4.6 & 0.844 & 0.994 \\  
        50 &  4.2 & 4.5 & 0.866 & 0.995 \\  
        \bottomrule
    \end{tabular}
    \caption{\textbf{Effect of the number of samples from the cascaded diffusion model.} The table reports metrics of the \textit{final} hand mesh on FreiHAND. In contrast to \cref{tab:samples_joint}, which evaluates the Stage-1 joint output, this measures the full cascaded pipeline.}
    
    \label{tab:multi_cascade}
\end{table}

\subsection{Multi-hypotheses}
For the FreiHAND dataset, we also analyze the impact of the number of samples on the performance of the cascaded diffusion model. Specifically, we generate 50 joint hypotheses from the joint diffusion model and feed them into the Mesh LDM. The corresponding quantitative results are presented in Table~\ref{tab:multi_cascade}. Similar to the joint diffusion model, as the number of generated samples increases, the best performance of cascaded diffusion model also improves.

While multi-sampling improves quantitative performance, the visual differences between generated samples are subtle. Notably, the variations among hypotheses primarily affect the hand's shape rather than its pose. This occurs because the joint diffusion model generates a joint hypothesis, which then conditions Mesh LDM. At this stage, the pose configuration is already determined, and Mesh LDM reconstructs the hand mesh based on the given joint sample. Additionally, the reconstructed mesh aligns with the input image while refining shape properties such as finger thickness.

\begin{figure*}[htb]
  \centering
   \includegraphics[width=0.95\linewidth]{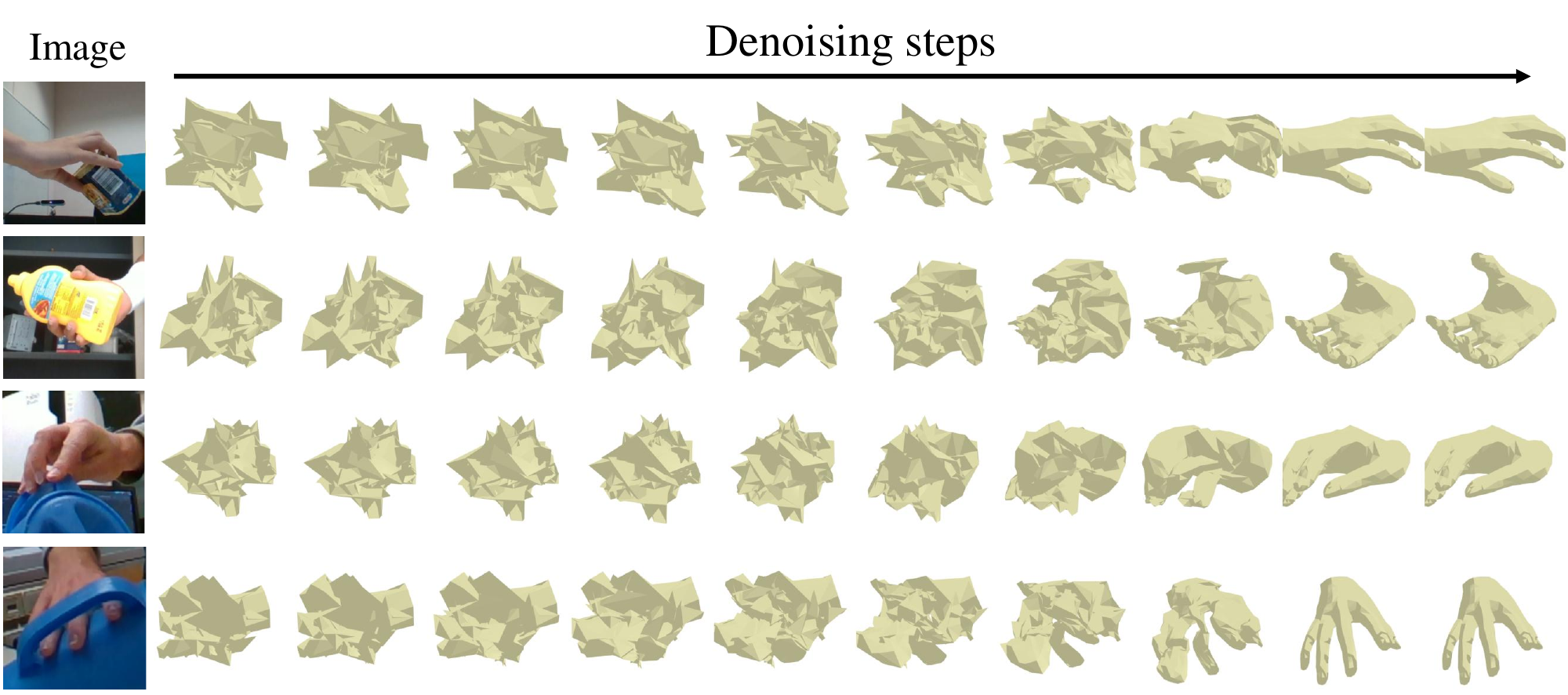}
   \caption{\textbf{DDIM denoising process} on HO3Dv2 with DDIM step 10.}
   \label{fig:denoising_ho3d}
\end{figure*}

\subsection{DDIM step}
Figure~\ref{fig:denoising_ho3d} visualizes the denoising process of a hand mesh during DDIM inference. Initially, the hand mesh exhibits minimal structural definition. As the DDIM process progresses, the hand pose becomes more articulated after half of the DDIM steps. After that surface details of hand mesh gradually emerge. This demonstrates that diffusion in the latent space effectively captures both pose and surface information throughout the denoising process.

\begin{figure*}[htb]
  \centering
   \includegraphics[width=\linewidth]{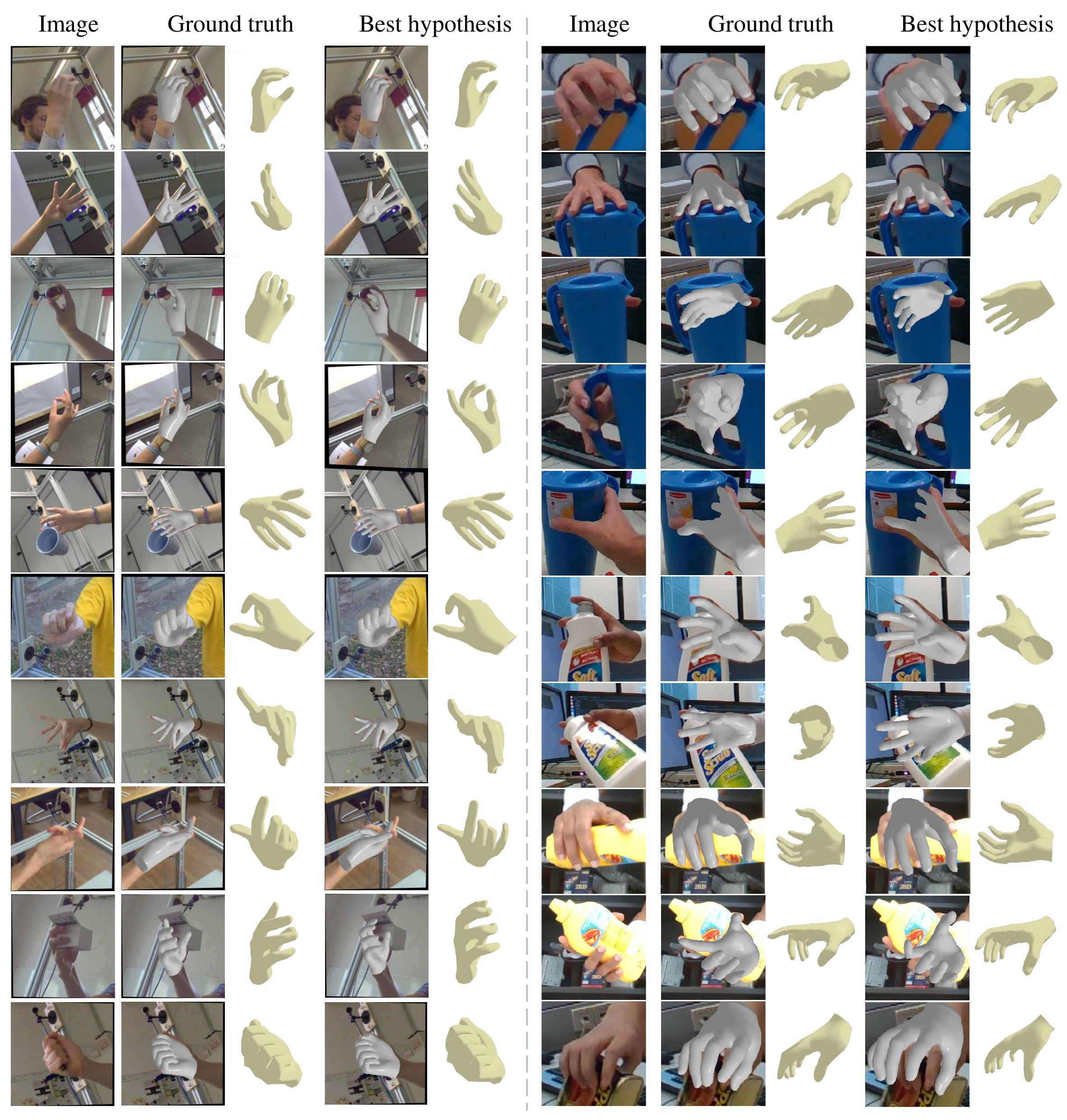}
   \caption{\textbf{Qualitative results} on FreiHAND and HO3Dv2 dataset.}
   \label{fig:multihypo_final}
\end{figure*}
\begin{table}[h]
\centering

\begin{tabular}{c|ccc|c|c}
\toprule
\small
& \multicolumn{3}{c|}{Ours} & \multirow{2}{*}{Hamba} & \multirow{2}{*}{WiLoR} \\
DDIM & 1 & 5 & 10 & & \\
\hline
Time (ms) & 40 & 140 & 260 & 40 & 50 \\
\bottomrule
\end{tabular}
\caption{\textbf{Comparison of inference time.} Inference time (in milliseconds) across different methods and ours different DDIM steps.}
\label{tab:inference_time}
\end{table}

\subsection{Inference speed}
Table~\ref{tab:inference_time} compares the inference times of our method (across different DDIM steps) with Hamba~\cite{dong2024hamba} and WiLoR~\cite{potamias2024wilor}, using a ResNet encoder. While the 10-step variant is slower, our 1-step variant matches the speed of existing methods. Furthermore, recent work on one-step distillation~\cite{salimans2022progressive} suggests that faster variants are feasible. Importantly, our model maintains robustness under occlusions, making it suitable for applications such as robotic grasping and hand–object interaction.

\end{document}